\documentclass[letterpaper]{article} 
\usepackage[submission]{aaai25}  
\usepackage{times}  
\usepackage{helvet}  
\usepackage{courier}  
\usepackage[hyphens]{url}  
\usepackage{graphicx} 
\usepackage{natbib}  
\usepackage{caption} 
\usepackage{amsmath} 
\usepackage{amssymb} 
\usepackage{multirow}
\usepackage{subfig}
\usepackage{graphicx}
\usepackage[table,xcdraw]{xcolor}
\usepackage{mathabx}
\usepackage{booktabs}
\usepackage{bm}
\usepackage{arydshln}

\usepackage{utfsym} 

\usepackage{color}

\usepackage{algorithm}
\usepackage{algorithmic}
\usepackage{subfiles}

\usepackage{newfloat}
\usepackage{listings}

\DeclareCaptionStyle{ruled}{labelfont=normalfont,labelsep=colon,strut=off} 
\floatstyle{ruled}
\newfloat{listing}{tb}{lst}{}
\floatname{listing}{Listing}
\title{Big Brother is Watching: Proactive Deepfake Detection via Learnable Hidden Face}

\author{
    Hongbo Li, Shangchao Yang, Ruiyang Xia, Lin Yuan, Xinbo Gao
}
\affiliations{
    \textsuperscript{\rm 1}Association for the Advancement of Artificial Intelligence\\

    1101 Pennsylvania Ave, NW Suite 300\\
    Washington, DC 20004 USA\\
    proceedings-questions@aaai.org
}

\usepackage{bibentry}

\begin{document}

\maketitle

%

\begin{abstract}
As deepfake technologies continue to advance, 
passive detection methods struggle to generalize 
with various forgery manipulations and datasets. 
Proactive defense techniques have been actively studied 
with the primary aim of preventing deepfake operation effectively working. 
In this paper, we aim to bridge the gap between passive detection and 
proactive defense, and seek to solve the detection problem 
utilizing a proactive methodology. 
Inspired by several watermarking-based forensic methods, 
we explore a novel detection framework based on the concept of 
``hiding a learnable face within a face''. 
Specifically, relying on a semi-fragile invertible steganography network, 
a secret template image is embedded into a host image imperceptibly, 
acting as an indicator monitoring for any malicious image forgery 
when being restored by the inverse steganography process. 
Instead of being manually specified, the secret template is optimized 
during training to resemble a neutral facial appearance, 
just like a ``big brother'' hidden in the image to be protected. 
By incorporating a self-blending mechanism and robustness learning strategy 
with a simulative transmission channel, a robust detector is built to accurately distinguish 
if the steganographic image is maliciously tampered or benignly processed. 
Finally, extensive experiments conducted on multiple datasets demonstrate 
the superiority of the proposed approach over competing passive and proactive 
detection methods. 
\end{abstract}
\section{Introduction}
\label{sec：intro}
As a representative application of recently advanced generative models, 
{\it deepfake} has emerged as a powerful tool to create fake face images 
or video, by modifying the identity, attributes, or expression of 
the original face. 
On the one hand, the deepfake technology has been used in 
film industry, entertainment apps, and advertisement, 
providing positive influence and benefits. 
Sadly, the ``double-edged sword'' nature of deepfake also makes it 
especially useful in various negative applications to create fake news, 
assist in cyberfraud, or produce pornographic films. 

To mitigate the negative impact of deepfake technologies, 
various defense methods have been developed, which can be roughly 
categorized into two types: passive detection and proactive defense. 
The former targets at detecting if an image has been forged primarily 
based on analyzing the visual content of the suspicious image solely. 
Despite their considerable accuracy, many methods still face limitations 
in generalizing across different forgery methods and datasets. 
In contrast with passive defense, proactive defense fights against 
deepfakes proactively by adding special perturbations or watermarks 
to the source images or video before they are released. 
When a malicious user attempts to use the published images for 
creating deepfakes, they may only obtain visually damaged results 
or leave visible cues for sourcing and forensic analysis. 

In this paper, we attempt to bridge the gap between passive detection 
and proactive defense, by proposing a novel framework to enhance the 
deepfake detection performance utilizing a proactively embedded visual signal. 
Inspired by several watermarking-based proactive defense methods
~\cite{asnani2022pimd,neekhara2024FaceSigns,Zhao2023pddiw,Wu2023sepmark}, 
we explore to embed a learnable image template into a face image via a
semi-fragile invertible steganography network. 
This makes sure the restored template sensitively reflect the potential 
usage of deepfakes applied on the steganographic image, meanwhile keeping 
immune to common image processing operations such as JPEG compression. 
Through a delicate training strategy, 
we optimize the learnable template to become a face as well, 
showing an neutral facial appearance without 
explicitly revealing any realistic identity. 
This also ensures that the steganographic image induces 
imperceptible changes compared to the original image. 
To summarize, we make the following contributions: 
\begin{itemize}
    \item A novel framework for proactive deepfake detection built on a fresh concept of 
    \textit{hiding a learnable face into a face} is proposed, which simulates a scenario where a 
    ``big brother'' hidden inside an image is ``watching'' over any malicious forgery attempt. 
    The central to the framework is an invertible image steganography architecture, 
    which proactively embeds a secret face template into a face image to be protected, 
    and makes the restored secret face effectively discriminate the authenticity 
    of the steganographic image after being transmitted in public channels.     
    \item A delicate training strategy is developed, which consists of a specialized transmission
    channel simulating both benign and malicious image manipulations, along with multi-task loss 
    functions for optimizing both deepfake detection and face template renewal. 
    \item Extensive experiments conducted in cross-dataset scenarios 
    demonstrate the superior performance of the proposed approach in 
    deepfake detection, image quality, and robustness against common image processing. 
\end{itemize}
\section{Related Work}
\label{sec:related_work}

\subsubsection{Passive Deepfake Detection} 
A considerable amount of research efforts have been devoted into 
the passive deepfake detection technology. 
Proposed approaches aim to build an effective deepfake classifier 
by analyzing intrinsic inconsistencies between real and fake samples 
in image pixel domain~\cite{Dong2023CADDM,Cao2022RECCE,Shiohara2022SBI}, 
frequency domain~\cite{qian2020F3Net,li2021FDFL}, 
and multi-modal domain~\cite{Haliassos2021Lips,Shao2023DGM}. 
Particularly, Shiohara et al.~\cite{Shiohara2022SBI} proposed a self-blended image (SBI) fuse strategy, 
which generates realistic forged images through self-keypoint transformations and data augmentation. 
This technique, independent of specific forgery methods, demonstrates excellent generalization 
capabilities to adapt to various types of forgeries, which is also employed in the proposed proactive detection approach. 

\subsubsection{Proactive Deepfake Defense}
Proactive defense strategies against deepfakes can be broadly classified into two main approaches: 
proactive attack methods~\cite{ruiz2020disrupt,huang2021initiative,huang2022CMUA,yang2021defending}, 
which disrupt the output of deepfake models, 
and proactive forensic methods, which focus on tracing ownership or verifying the authenticity of manipulated images. 
Our method is highly related to the latter group, reviewed as follows: 
\cite{Asnani2023MaLP}
introduced a dual-branch framework that learns both local and global features to generate a deepfake prediction map.
\cite{Hu2023DRAW}
proposed a method to protect RAW data captured by cameras.
Even after processing through the Image Signal Processing (ISP) pipeline and undergoing common transformations like blurring or compression, DRAW’s localization network can still identify manipulated regions.
Recently,
\cite{zhang2024editguard}
introduced a framework that achieves both copyright protection and tampering localization by embedding semantic bit information and a predefined solid image. 
They found that when a watermark image embedded using an invertible neural network undergoes certain forgeries, the damaged area of the restored image closely aligns with the forged area, making it possible for deepfake localization.

In the realm of proactive forgery binary detection, 
\cite{asnani2022pimd}
applied steganography by training a set of orthogonal templates and randomly selecting one to embed secure information into target images.
Deepfake is detected by extracting the template from the target image and comparing its similarity to the original template.
\cite{yu2021artificial,sun2022faketracer}
embedded watermark information into facial images, ensuring that when deepfake models use these watermarked images for training, the watermarks are carried over to the generated deepfakes, making them identifiable.
\cite{Wang2021FakeTagger}
introduced a method that embeds a robust watermark into images, which remains intact against forgery attempts, thus allowing the tracing of image ownership back to the original uploader.
\cite{Zhao2023pddiw,neekhara2024FaceSigns}
proposed a proactive defense method based on semi-robust identity watermarks, which are resistant to deepfake algorithms and can withstand common image transformations. 
This allows for authenticity verification by computing the correlation between extracted watermarks and the original.
More recently, 
\cite{Wu2023sepmark,Liu2023BiPro}
introduced deep separable watermarks, a new paradigm in proactive forensics against deepfakes. This framework includes an encoder and two decoders with two robustness characteristics. 
With a single watermark embedding, the robust and semi-robust decoders can extract the watermark separately
at different levels of robustness, aiding both deepfake tracing and detection.

\begin{figure*}[t]
    \centering
     \includegraphics[width=1\textwidth]{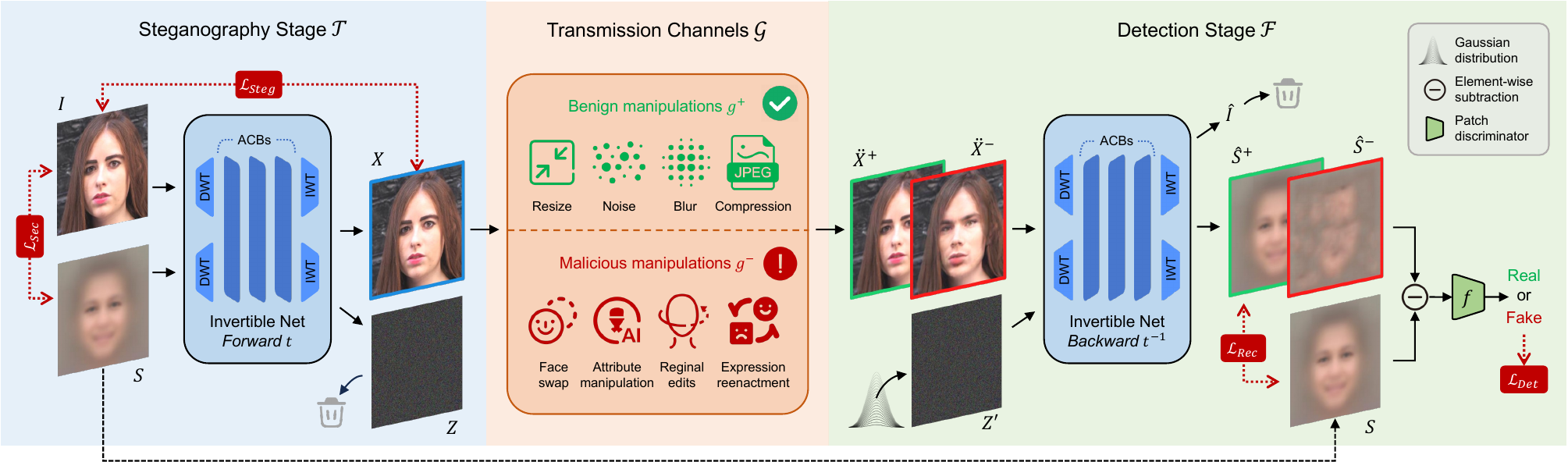}
     \caption{The overall framework of proposed proactive deepfake detection approach.}
     \label{fig:framework}
\end{figure*}

\section{Method}

\subsection{Problem Formulation}
Given $I$ indicating a face image to be protected, 
our proactive detection scheme starts with a steganography module $\mathcal{T}_\phi$ 
(with $\phi$ indicating module parameters), 
which hides a learnable secret template $S$ into the input image $I$: 
\begin{equation}
    \mathcal{T}_\phi\left(I, S\right) \cong I, 
\end{equation}
where the resulted steganographic image $\mathcal{T}_\phi\left(I, S\right)$ 
should closely resemble the original input $I$. 
The steganographic images are then released to public transmission channels, 
where they undergo various image manipulations including not only commonly used 
processing such as image compression but also malicious ones such as deepfakes. 
We consider the former as benign manipulations (denoted as $\mathcal{G}^+$) 
and any image processed by them $\mathcal{G}^+(\mathcal{T}_\phi\left(I, S\right))$ as real. 
Conversely, the latter malicious manipulations are denoted as $\mathcal{G}^-$ 
and the corresponding processed image $\mathcal{G}^-(\mathcal{T}_\phi\left(I, S\right))$ is considered as fake. 
In the forensic stage, we expect to build a deepfake detection module $\mathcal{F}_\theta$ 
(with parameters $\theta$) capable of predicting the authenticity of the manipulated image 
with the help of the pre-embedded template $S$ (assuming the analyzer has access to $S$). 
Hence, the training objective for our proactive detection is formulated as
\begin{multline}
    \min_{\phi, \theta, S} \Big\{ -\mathbb{E}_{I, *\in\{+,-\}} \left[ y^* \log\big( \mathcal{F}_\theta(\mathcal{G}^*(\mathcal{T}_\phi(I, S), S) \big) \right. + \\\left. (1-y^*) \log\big( 1 - \mathcal{F}_\theta(\mathcal{G}^*(\mathcal{T}_\phi(I, S), S) \big) \right] \Big\},
\end{multline}
where $y^*$ refers to the authenticity label of manipulated image $\mathcal{G}^*(\mathcal{T}_\phi\left(I, S\right))$. 

\subsection{Motivations}\label{sec:motivation}
The designing of the proposed approach 
is first inspired by the watermarking-based deepfake forensic methods
~\cite{asnani2022pimd,neekhara2024FaceSigns,Zhao2023pddiw,Wu2023sepmark} 
where certain watermarks are embedded into an image such that any forgery 
applied will impact the integrity of watermark extraction. 
The detection model can infer whether the watermarked image is forged based on 
analyzing the difference between the restored and embedded watermarks. 
However, most methods within this type struggle with 
achieving a good balance between image imperceptibility, 
detection accuracy, and watermarking robustness. 

We further observe that the invertible neural network (INN) 
shows remarkable steganography performance in hiding high-capacity visual data 
(e.g., image or even video) into a cover image without impacting imperceptibility much. 
Yet, the INN-based steganography is quite sensitive to semantic image manipulations 
- any manipulation in the steganographic domain can sensitively affect the results 
of the restoration domain. This means, similar manipulation traces happening on the 
steganographic image can be reflected on the restored secret image as well. 
This effect has been verified by~\cite{deng2023pirnet,zhang2024editguard}, 
proven effective in tasks of tamper localization, copyright protection, 
and privacy-preserving image restoration. 

Inspired by the above findings, 
we aim to explore the most suitable steganography medium and 
training strategy based on INN for proactive deepfake detection. 
Our hypothesis is that the secret data being hidden in the 
cover image should be learned instead of manually specified 
as most existing approaches do. Thus the secret hidden medium 
can be better adapted with the deepfake detection task. 
The steganography network should be semi-fragile: The restored 
secret data should well resemble the originally embedded on if 
the steganographic image has not undergo any manipulation or 
only been processed by common processing tools like JPEG compression. 
Oppositely, if the steganographic image has been maliciously manipulated 
by deepfakes, the restored data should sensitively differ from the original, 
reflecting the traces of deepfakes. 

\subsection{Framework Design}

Central to our design is a novel concept of {\it hiding a learnable face within a face}. 
This approach allows the embedded face to act as an hidden ``inspector,'' 
monitoring for any tampering or forgery attempts on the host face. 
As discussed in Section~\ref{sec:motivation}, we employ an invertible 
neural network (INN) for image steganography. The steganographic image 
closely resembles the original, but the hidden face restored by the INN 
can sensitively reveal traces of deepfakes as applied on the steganographic image. 
Common image processing techniques, such as filtering and compression, 
should not significantly affect the integrity of the restored face. 
By analyzing the differences between the original and restored hidden face, 
an effective deepfake detector can be developed. 
Notably, the secret image is not manually specified; 
instead, it is optimized to approximate the entire set of input face images 
during the training process of the deepfake detector. 
This optimization results in a secret image showing a natural facial appearance 
with a similar spatial structure to the cover image, facilitating imperceptible steganography. 
The overall framework is illustrated in Figure~\ref{fig:framework}. 
The key components of the proposed framework are described in detail as follows. 

\subsubsection{Invertible Steganography Module} 
We employ the same INN structure as described in~\cite{Jing2021hinet} 
for our image steganography module. The forward pass of image steganography, 
denoted as $t$, starts with converting the input image $I$ 
and secret image $S$ into wavelet sub-bands 
with the discrete wavelet transformation (DWT).  
The wavelet feature maps are then fed into an 
invertible convolutional network features by 
a sequence of affine coupling blocks (ACBs). 
The output wavelet sub-bands nonlinearly transformed by ACBs 
are converted back to image domain with the inverse wavelet transformation (IWT).  
This results in the steganographic image $X=t\left( I, S \right)$ 
that is expect to resemble the original input $I$, 
yielding a steganography loss function constrained by pixel-level L2 distance: 
\begin{equation}
    \mathcal{L}_{Steg}= \mathbb{E}_I\| I - t\left( I, S \right) \|_2.
\end{equation}
Due to the nature of INN, the steganography process 
also produces a redundant image $Z$ that does not need to be preserved. 
Following~\cite{Jing2021hinet}, a noise image $Z'$ randomly sampled 
from a Gaussian distribution is utilized to substitute $Z$ 
during the restoration stage, which is described in Section~\ref{sec:restoration}. 
We incorporate only four ACBs in our steganography model, 
which has been proven to provide effective hiding performance 
while keeping the model lightweight~\cite{yuan2024profaces}.

\begin{figure*}[t]
     \centering
     \includegraphics[width=0.9\textwidth]{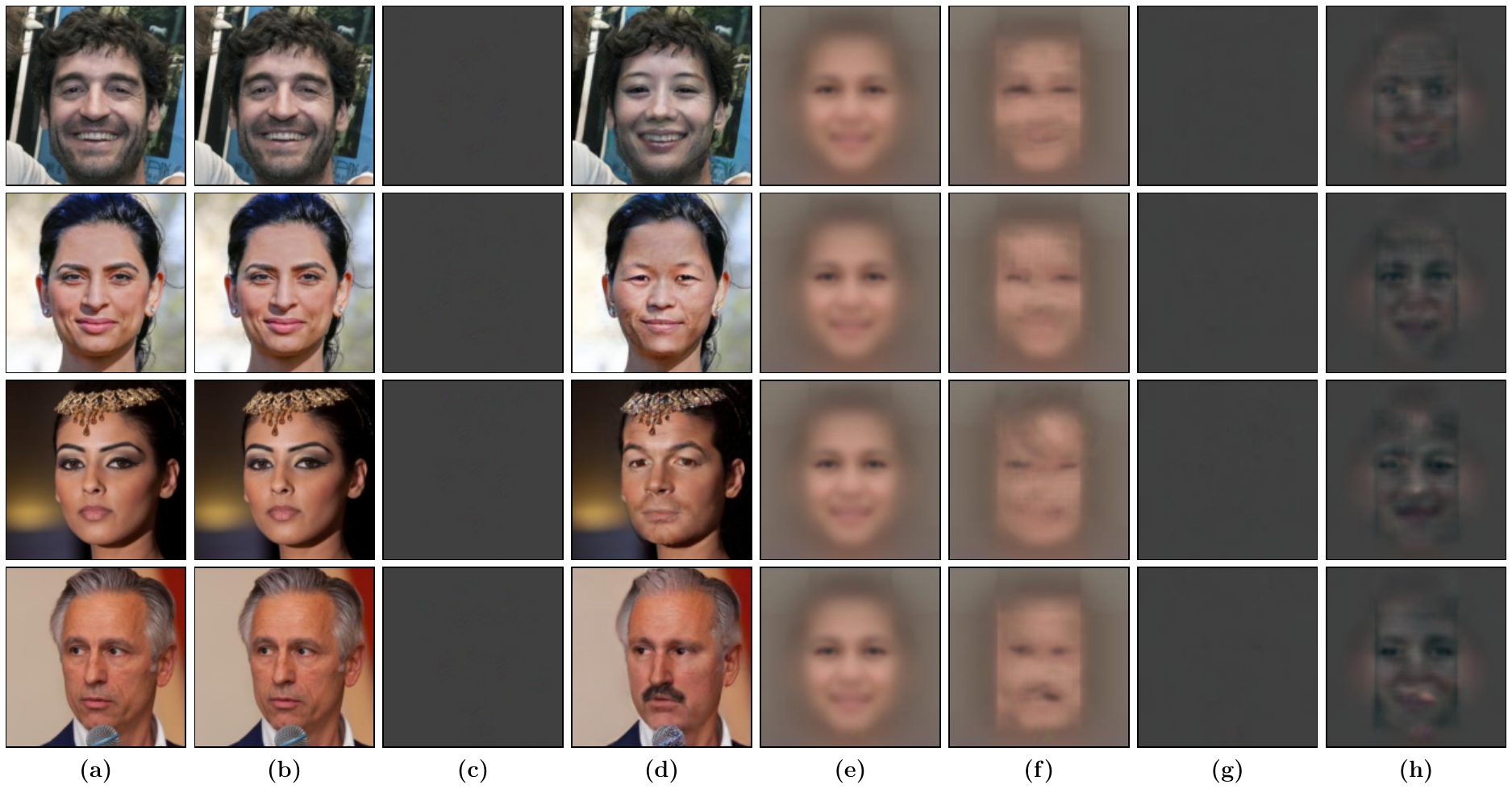}
     \caption{
      Image samples within the framework: 
     (a) Original image; 
     (b) Steganographic image; 
     (c) Residual between (a) and (b); 
     (d) Steganographic image maliciously manipulated by SimSwap~\cite{Chen2020simswap}; 
     (e) Template restored from benignly manipulated steganographic image; 
     (f) Template restored from maliciously manipulated steganographic image; 
     (g) Residual between (e) and the original template;
     (h) Residual between (f) and the original template. 
     All above residual images are amplified by a nonlinear square operation for more visible display. 
     }
     \label{fig:visualization_show_all}
\end{figure*}

\subsubsection{Image Manipulations in Transmission Channels} 
The steganographic image is intended to be publicly available for 
used in transmission channels, where it may undergo various manipulations—both benign, 
like filtering and compression, and malicious, such as deepfakes. 
We expect the detector effectively sense any malicious manipulation 
without treating the benign manipulation as fake. 
Hence, we simulate a transmission channel in training by applying 
benign and malicious manipulations randomly on the steganographic image. 
In our experiments, we consider four commonly used manipulations as benign: 
Gaussian blurring, Gaussian noising, rescaling, and JPEG compression. 
To simulate malicious forgery, we employ the self-blending mechanism~\cite{Shiohara2022SBI}, 
which has proven effective in generalizing 
various deepfakes without requiring labeled samples in training. 
Specifically, we randomly create self-blended images (SBIs) 
from input real images, and treat the SBIs as fake samples. 
Thus, for each input image, we create two transmitted versions serving 
as real and fake samples respectively for training the deepfake detector. 
The real and fake samples corresponding to a steganographic image $X$ 
are formulated as $\ddot{X}^{+} = g^+\left(X\right)$ and 
$\ddot{X}^{-} = g^+\left(g^-\left( X \right)\right)$ respectively, 
where $g^+$ refers to a random benign operation or leaving the image unchanged, 
and $g^-$ denotes the self-blending operation as defined in~\cite{Shiohara2022SBI}. 
To enhanced to the detection robustness, we also apply random benign manipulations 
on synthesized fake SBI samples. 

\subsubsection{Secret Image Restoration and Deepfake Detection} 
\label{sec:restoration}
Employing the same invertible steganography network, 
we attempt to restore the hidden template $\hat{S}$ from the transmitted image: 
$\hat{S} = t^{-1}\left(\ddot{X}^*, Z'\right), * \in\{+,-\}$, 
with the help of a noise variable $Z'$ randomly 
sampled from a Gaussian distribution, substituting the lost variable $Z$ during the steganography stage. 
Here, $t^{-1}$ refers to the backward process of the steganography module, 
composed of DWT, inverse ACBs, and IWT operations sequentially. 
We then employ a simple patch discriminator~\cite{Isola2017patchGAN} as the deepfake classifier model, 
denoted as $f$, which takes as input the residual between the restored template 
and the original one $\Delta S = S - \hat{S}$. 
The deepfake classifier is optimized on the binary cross-entropy loss 
$\mathcal{L}_{Det}$ as follows:
\begin{multline}
    \mathcal{L}_{Det} = - \mathbb{E}_{(\Delta S, y)} \left[y \cdot \log\left( f\left(\Delta S\right) \right) + \right.\\
    \left.(1-y)\cdot \log\left( 1 - f\left(\Delta S\right)\right) \right],
\end{multline}
where $y$ refers to the class label corresponding to $\Delta S$. 

\subsubsection{Secret Image Optimization}
To facilitate the proactive detection performance, 
we utilize two additional loss functions to optimize 
the secret template $S$. 
The goal is to train a single template, $S$, 
which is designed to approximate every input 
face image during the training process.
\begin{equation}
    \mathcal{L}_{Sec} = \mathbb{E}_{I} \| \mathrm{DWT}_{LL}\left(S\right) - \mathrm{DWT}_{LL}\left(I\right)\|_2,
\end{equation}
where $\mathrm{DWT}_{LL}$ denotes the LL sub-band of DWT. 
This key design produces a final secret template that resembles the average of all input faces, 
resulting in a natural and realistic human facial appearance. 
Additionally, we optimize the restored secret template 
to approximate the original one when real samples 
$\ddot{X}^{+}$ are presented to the restoration process: 
\begin{equation}
     \mathcal{L}_{Rec} = \mathbb{E}_{(\ddot{X}^{+}, Z'\sim\mathcal{N}(0,1))} \| t^{-1}\left(\ddot{X}^{+}, Z'\right) - S\|_2.
\end{equation}
The loss function helps the model further differentiate the 
restored template corresponding to real and fake samples.

\begin{table*}[t]
\centering
\footnotesize
\tabcolsep=0.4em
\begin{tabular}{cccccccccc}
\toprule
\multirow{2}{*}{Dataset}   & \multirow{2}{*}{Method} & \multicolumn{4}{c}{Identity Swapping}   & \multicolumn{2}{c}{Attributes Editting} & \multicolumn{1}{c}{Anonymization}         &  \multirow{2}{*}{AVG} \\ \cmidrule(lr){3-6} \cmidrule(lr){7-8} \cmidrule(lr){9-9} 
                           &                         & SimSwap & FaceShifter & FaceSwap & MFaceSwap & StarGAN2 & TTedit & FIT  \\ \midrule \midrule
\multirow{7}{*}{FFHQ}      & RECCE                   &    33.56     &      36.46       &     46.76     &       70.82         &     13.56     &  80.17      &  63.19   &   49.22   \\
                           & SBI                     &     72.60    &    77.01        &      83.44    &    97.36            &     65.11     &   83.50     &  58.91   &      76.85\\
                           & CADDM                   &     68.28    &   80.65          &      89.39    &      89.88           &    44.00    &  75.72   &   64.69 & 73.23   \\
                           \cmidrule{2-10}
                           & PDDIW                  &     96.88    &    98.96      &    63.14    &     86.13           &     85.43     &   54.15     &   99.06  & 83.39    \\
                           & SepMark                &     100.00    &    50.00         &    50.00      &       50.00         &     100.00    &     97.47   &  94.33   &   77.40   \\
                           & FaceSigns            &     99.65    &       99.90      &    99.96      &     99.75           &      100.00   &  \textbf{99.90}      & 99.92    &  99.87   \\
                           \cmidrule{2-10}
                           & Ours                    &   \textbf{100.00}     &       \textbf{100.00}     &    \textbf{99.99}      &       \textbf{99.84}        &    \textbf{100.00}      &     99.23   &   \textbf{100.00}  &  \textbf{99.87}    \\ \midrule
\multirow{7}{*}{CelebA-HQ} & RECCE                   &    69.25   &       71.47     &    75.09    &     96.96           &    34.73      &   70.69     &   94.72  &   73.27   \\
                           & SBI                     &    79.06     &     71.05        &    82.60      &    99.11            &  60.12      &   86.66     &   60.20  &   78.40   \\
                           & CADDM                   &   70.01     &   85.30          &    92.43      &       95.57        &   44.02       &   85.72     &  74.12   &  78.17    \\
                           \cmidrule{2-10}
                           & PDDIW                  &     97.80    &       98.63      &    69.11      &     93.03           &     85.11     &    62.05   &  98.66   &   86.34   \\
                           & SepMark                &     99.97   &    50.00        &     50.00     &     50.00          &   99.97       &    \textbf{99.97}    &  96.67   &   78.08   \\
                           & FaceSigns            &    99.58     &      99.85       &   99.96       &       99.80         &     100.00    &     99.53   &    100.00  &   99.82  
                           \\
                           \cmidrule{2-10}
                           & Ours                    &   \textbf{100.00}    &   \textbf{100.00}         &    \textbf{99.99}      &     \textbf{99.92}           &   \textbf{100.00}       &     99.95  &  \textbf{100.00}   &   \textbf{99.98}   \\ \midrule
\multirow{7}{*}{VGGFace2}  & RECCE                   &   48.02      &   54.08          &   54.99       &    67.59           &   21.73       &   76.10     &  84.01   &   58.07   \\
                           & SBI                     &    87.36    &     89.51        &    96.11      &       98.07         &    59.81      &  88.33     &  85.52   &   90.67   \\
                           & CADDM                   &    81.65     &     79.73        &     98.09     &      91.55         &     76.47    &   82.97     &  85.94   &   85.20   \\
                           \cmidrule{2-10}
                           & PDDIW                  &    49.83     &      49.41       &     49.96     &      49.39          &    48.83      &  49.87     &  47.86  &  49.31  \\
                           & SepMark                &    99.03     &       50.00     &   50.00       &      50.00          &    99.93      &    99.56    &  67.12   &  73.66   \\
                           & FaceSigns            &     97.20    &     99.52        &    99.96      &      99.75          &     100.00     &   100.00     &  95.70   &  98.88  
                           \\
                           \cmidrule{2-10}
                           & Ours                    &    \textbf{100.00}     &    \textbf{100.00}         &     \textbf{100.00}     &      \textbf{100.00}          &  \textbf{100.00}        &   \textbf{100.00}     &   \textbf{100.00}  &   \textbf{100.00}   \\ 
                           \bottomrule
\end{tabular}%
\caption{Deepfake detection performance in terms of AUC (\%) 
for various deepfake techniques on different datasets. 
RECCE \cite{Cao2022RECCE}, SBI \cite{Shiohara2022SBI}, and CADDM \cite{Dong2023CADDM} 
are passive detection methods whereas FaceSigns~\cite{neekhara2024FaceSigns}, 
SepMark~\cite{Wu2023sepmark}, and PDDIW~\cite{Zhao2023pddiw} are proactive methods.}
\label{tab:compare_eval_1}
\end{table*}

\subsubsection{Final Loss Objective}
To summarize, the final loss function for training is formulated as
\begin{equation}
    \mathcal{L}_{Total}=
    \lambda_{1} \cdot \mathcal{L}_{Steg} + 
    \lambda_{2} \cdot \mathcal{L}_{Sec} +
    \lambda_{3} \cdot \mathcal{L}_{Rec} +
    \lambda_{4} \cdot \mathcal{L}_{Det},
\end{equation}
where $\lambda_{i}$ are the weights balancing different loss terms.

\section{Experiment}

\begin{table}[t]
\centering
\tabcolsep=0.3em
\begin{tabular}{c}

    \begin{minipage}{\linewidth}
        \centering
        \resizebox{\linewidth}{!}{%
        \begin{tabular}{lccccccc}
        \toprule
        \multirow{2}{*}{\begin{tabular}[c]{@{}c@{}}\bf JPEG\\ \bf (QF=75)\end{tabular}} & \multicolumn{4}{c}{Identity Swap} & \multicolumn{2}{c}{Attribute Edit} & FA \\ \cmidrule(lr){2-5} \cmidrule(lr){6-7} \cmidrule(lr){8-8} 
                                & SS & FST & FS & MFS & SG2 & TTE & FIT \\
        \midrule 
        PDDIW & 97.82 & 98.58 & 69.18& 93.15& 85.10& 61.97& 98.67 \\
        SepMark & 99.07 & 50.00 & 50.00 & 50.04 & 100.00 & 100.00 & 89.17 \\
        FaceSigns & 99.65 & 99.78 & 99.94 & 99.70 & 100.00 & 99.70 &  100.00 \\
        \midrule
        Ours & \textbf{100.0} & \textbf{100.0} & \textbf{99.99} & \textbf{99.96} &  \textbf{100.0} &  \textbf{100.0} & \textbf{100.0} \\
        \bottomrule
        \end{tabular}
        }
    \end{minipage}%
    
    \\
    \addlinespace 

    \begin{minipage}{\linewidth} 
        \centering
        \resizebox{\linewidth}{!}{%
        \begin{tabular}{lccccccc}
        \toprule
         \multirow{2}{*}{\begin{tabular}[c]{@{}c@{}}\bf Noising \\ \textbf{(\bm{$\sigma$}=0.03)} \end{tabular}}   & \multicolumn{4}{c}{Identity Swap}  & \multicolumn{2}{c}{Attribute Edit} & FA \\ \cmidrule(lr){2-5} \cmidrule(lr){6-7} \cmidrule(lr){8-8} 
                                & SS & FST & FS & MFS & SG2 & TTE & FIT \\
        \midrule
        PDDIW & 97.87 & 98.65 & 69.79 & 93.36 & 86.70 & 64.57 & 98.73 \\
        SepMark & 99.07 & 50.00 & 50.00 & 50.00 & 100.00 & \textbf{100.00} & 88.37 \\
        FaceSigns & 95.30 & 99.83 & 99.98 & 99.77 & 100.00 & 99.77 & 100.00 \\
        \midrule
        Ours & \textbf{100.0} & \textbf{100.0} & \textbf{99.98} & \textbf{99.95} & \textbf{100.0} & 99.95 & \textbf{100.0} \\
        \bottomrule
        \end{tabular}
        }
    \end{minipage}

    \\
    \addlinespace 
    
    \begin{minipage}{\linewidth} 
        \centering
        \resizebox{\linewidth}{!}{
        \begin{tabular}{lccccccc}
        \toprule
        \multirow{2}{*}{\begin{tabular}[c]{@{}c@{}}\bf Rescaling\\ \bf (Factor=0.7)\end{tabular}} & \multicolumn{4}{c}{Identity Swap} & \multicolumn{2}{c}{Attribute Edit} & FA \\ \cmidrule(lr){2-5} \cmidrule(lr){6-7} \cmidrule(lr){8-8} 
                                & SS & FST & FS & MFS & SG2 & TTE & FIT \\
        \midrule
        PDDIW & 97.57 & 98.50 & 68.65 & 92.55 & 86.13 & 61.93 & 98.57 \\
        SepMark & 82.35 & 82.35 & 74.82 & 79.21 & 81.30 & 81.30 & 82.02 \\
        FaceSigns & 99.53 & 95.55 & 95.67 & 95.48 & 95.68 & 95.48 & 95.68 \\
        \midrule
        Ours & \textbf{100.0} & \textbf{100.0} & \textbf{100.0} & \textbf{99.97} & \textbf{100.0} & \textbf{99.81} & \textbf{100.0} \\
        \bottomrule
        \end{tabular}
        }
    \end{minipage}%
    
    \\
    \addlinespace 

    \begin{minipage}{\linewidth} 
        \centering
        \resizebox{\linewidth}{!}{%
        \begin{tabular}{lccccccc}
        \toprule 
        \multirow{2}{*}{\begin{tabular}[c]{@{}c@{}}\bf Blur \\ \textbf{(\bm{$\sigma$}=1.5)} \end{tabular}} & \multicolumn{4}{c}{Identity Swap} & \multicolumn{2}{c}{Attribute Edit} & FA \\ \cmidrule(lr){2-5} \cmidrule(lr){6-7} \cmidrule(lr){8-8} 
                                & SS & FST & FS & MFS & SG2 & TTE & FIT \\
        \midrule
        PDDIW & 97.63 & 98.57 & 69.04 & 93.23 & 85.75 & 61.37 & 98.67 \\
        SepMark & 99.93 & 50.00 & 50.00 & 50.00 & 100.00 & \textbf{100.00} & 94.15 \\
        FaceSigns & 99.70 & 99.87 & 99.96 & 99.83 & 100.00 & 99.83 & 100.00 \\
        \midrule
        Ours & \textbf{100.0} & \textbf{100.0} & \textbf{99.97} & \textbf{99.98} & \textbf{100.0} & 99.91 & \textbf{100.0} \\
        \bottomrule
        \end{tabular}
        }
    \end{minipage}

\end{tabular}
\caption{Robustness evaluation with different benign transformations on CelebA-HQ. SS, FST, FS, MFS, SG2, TTE, and FA refer to SimSwap, FaceShifter, FaceSwap, MobileFaceSwap, StarGAN2, TTedit, and face anonymization respectively.}
\label{tab:four-subtables}
\end{table}

\subsection{Experimental Settings}
\subsubsection{Dataset}
We primarily employ FFHQ~\cite{Karras2019ffhq} for training, 
which is split into three parts: 
30K images for training, 3K for validation, and 3K for testing. 
Additionally, we use VGGFace2 ~\cite{cao2018vggface2} and CelebA-HQ~\cite{Karras2017celebahq} 
in testing, simulating cross-dataset detection scenarios. 
We randomly sampled 3K images from each of the two datasets. 
All images in the experiments are scaled or cropped to $256\times256$. 

\subsubsection{Malicious Manipulations}
We evaluate the detection performance against multiple 
malicious image manipulations, including four face swapping algorithms 
(SimSwap~\cite{Chen2020simswap}, FaceShifter~\cite{Li2019faceshifter}, 
FaceSwap~\cite{wu2020faceswap}, and MFaceSwap~\cite{xu2022mobilefaceswap}), 
two facial attribute editing methods  (StarGAN2~\cite{Choi2020starganv2} 
and TTedit~\cite{jiang2021tte}), and a typical face anonymization method, FIT~\cite{gu2019fit}. 

\subsubsection{Training Details} 
The input batch size is set to 4, 
which results in a total 8 samples per batch for training the deepfake detector: 
4 negative samples generated by SBI~\cite{Shiohara2022SBI} and the other 
4 positive samples processed by benign manipulations at random. 
We randomly apply benign manipulations at various strengths during training. 
For JPEG, we simulate differentiable JPEG as proposed by~\cite{Reich2024diffJPEG}, 
and vary its quality factors between 50 and 95. 
For Gaussian blur, we fix the kernel size to 3 and 
uniformly vary its standard deviation in $[1,2]$ at random. 
For Gaussian noise, we uniformly sample its standard deviation in $[0, 0.05]$ at random. 
For image rescaling, we downsample the image according to a scaling factor uniformly sampled in 
$[0.5, 1.0]$ and upsample it back to the original size. 
The hyperparameters $\lambda_{1}$, $\lambda_{2}$, $\lambda_{3}$ and $\lambda_{4}$ 
are set to 2.0, 0.3, 1.0, and 8.0 respectively, which yield the optimial performance during validation. 
The Adam optimizers with learning rates of $10^{-4.5}$, $0.001$, and $0.0001$ are used for 
optimizing the steganography network, the deepfake discriminator, 
and the learnable face template respectively. 
The visualization of multiple images within the framework, as shown in Figure~\ref{fig:visualization_show_all}, 
clearly illustrates the differences in the residual template image between real and fake samples.

\subsubsection{Evaluation Baselines}
We compare with both proactive and passive literature methods. 
The proactive baselines include FaceSigns~\cite{neekhara2024FaceSigns}, 
SepMark~\cite{Wu2023sepmark}, and PDDIW~\cite{Zhao2023pddiw}, 
all with deepfake detection achieved via pre-embedded watermarks. 
We also include several competing passive detection methods in comparison, 
including RECCE~\cite{Cao2022RECCE}, SBI~\cite{Shiohara2022SBI}, and CADDM~\cite{Dong2023CADDM}. 
This is because the proposed method essentially relies on a passive detection 
stage to infer the image authenticity. It is therefore necessary to verify if 
the proposed method can compete those passive methods by using additionally 
embedded secret data.


\subsubsection{Evaluation Metrics}
We primarily utilize AUC 
to evaluate the deepfake detection performance. 
For the steganographic image quality, 
we compute the Peak Signal-to-Noise Ratio (PSNR), 
Structural Similarity Index Measure (SSIM)~\cite{Wang2004ssim},  
and Learned Perceptual Image Patch Similarity (LPIPS)~\cite{Zhang2018lpips} 
between the steganographic image and its original. 


\subsection{Deepfake Detection Evaluation}
We first evaluate the deepfake detection performance of the proposed approach 
in the condition without benign manipulations applied on the steganographic image. 
In this case, testing images are first protected by the proposed proactive 
steganography mechanism, resulting in real (positive) testing samples. 
The steganographic images are further processed by deepfake operations, 
resulting in fake (negative) testing samples.
For passive detectors, we test with forged images generated from 
the original images without steganography applied. 
As the results in Table~\ref{tab:compare_eval_1} show, 
the proposed approach outperforms both passive and proactive 
competitors for most datasets and deepfake methods. 
Passive methods primarily aim to develop a versatile detector 
and therefore struggle to generalize to various deepfake techniques. 
SepMark~\cite{Wu2023sepmark} and PDDIW~\cite{Zhao2023pddiw} 
incorporate only specific deepfake models in training, 
and therefore pose difficulties in coping with unseen deepfake techniques. 
Moreover, since most watermarking-based methods 
infers the authenticity by relying on the watermark 
bit recovery accuracy (BRA), selecting a proper BRA threshold 
suitable to all deepfake methods is hardly possible. 
In contrast, the proposed approach utilizes the self-blending 
mechanism~\cite{Shiohara2022SBI} in training to synthesize 
rich negative samples, therefore resulting in the optimal 
deepfake detection performance over diverse deepfake operations and datasets.

\begin{figure}[t]
  \centering
  \subfloat[JPEG]{\includegraphics[width=0.22\textwidth]{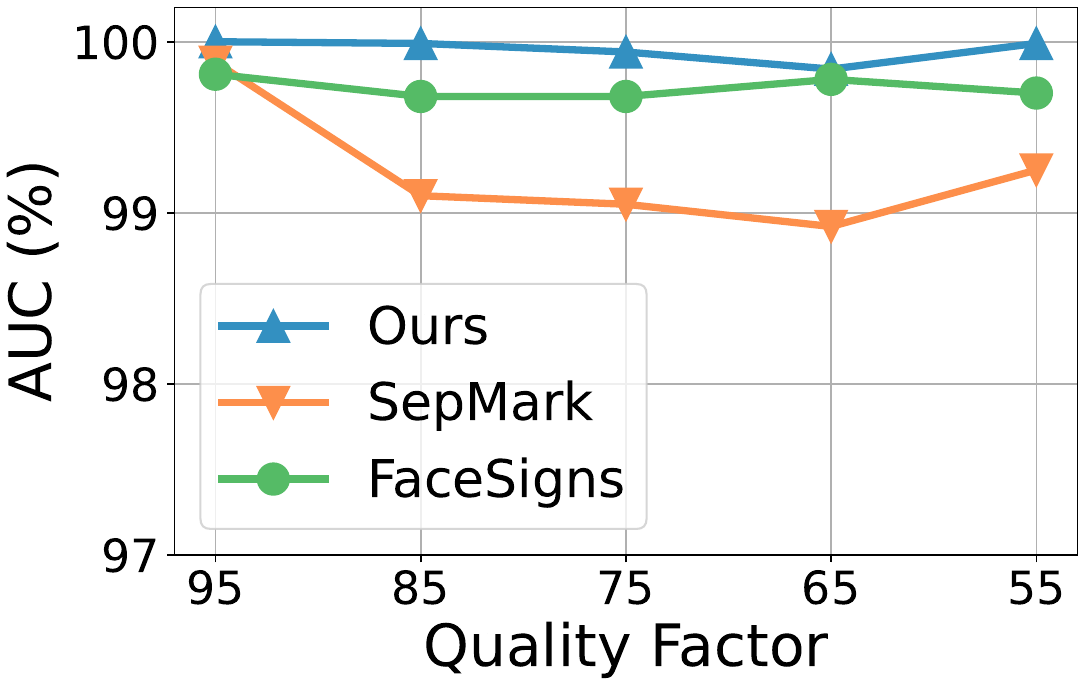}}
  \hfil
  \subfloat[Gaussian Noise]{\includegraphics[width=0.22\textwidth]{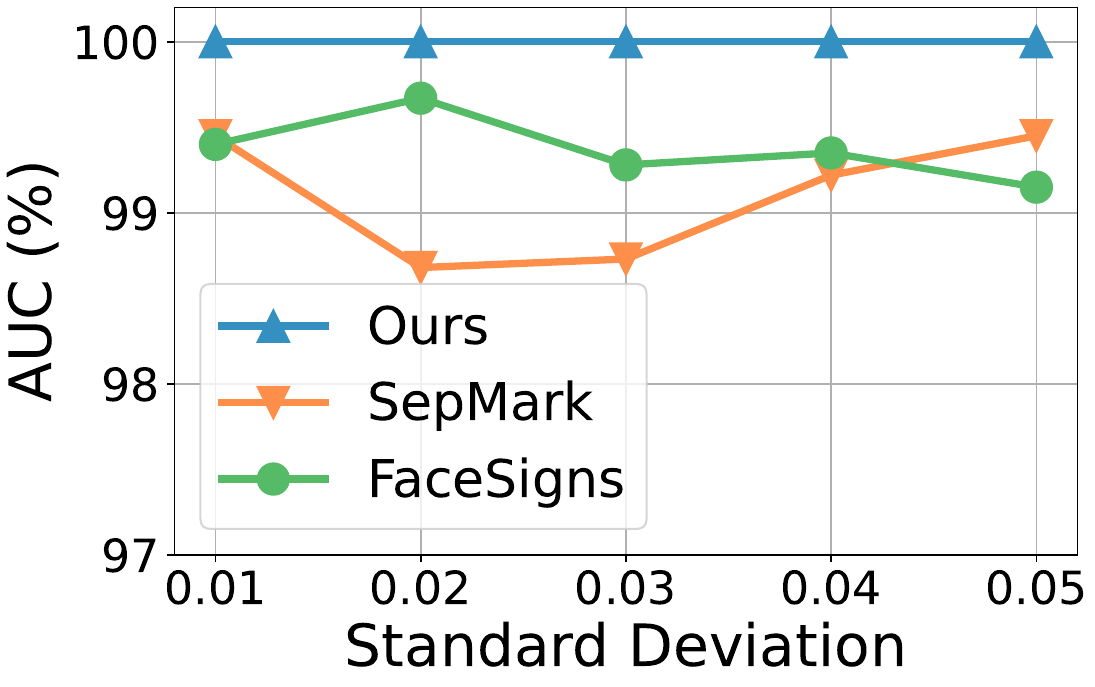}}
  \\
  \subfloat[Resize]{\includegraphics[width=0.22\textwidth]{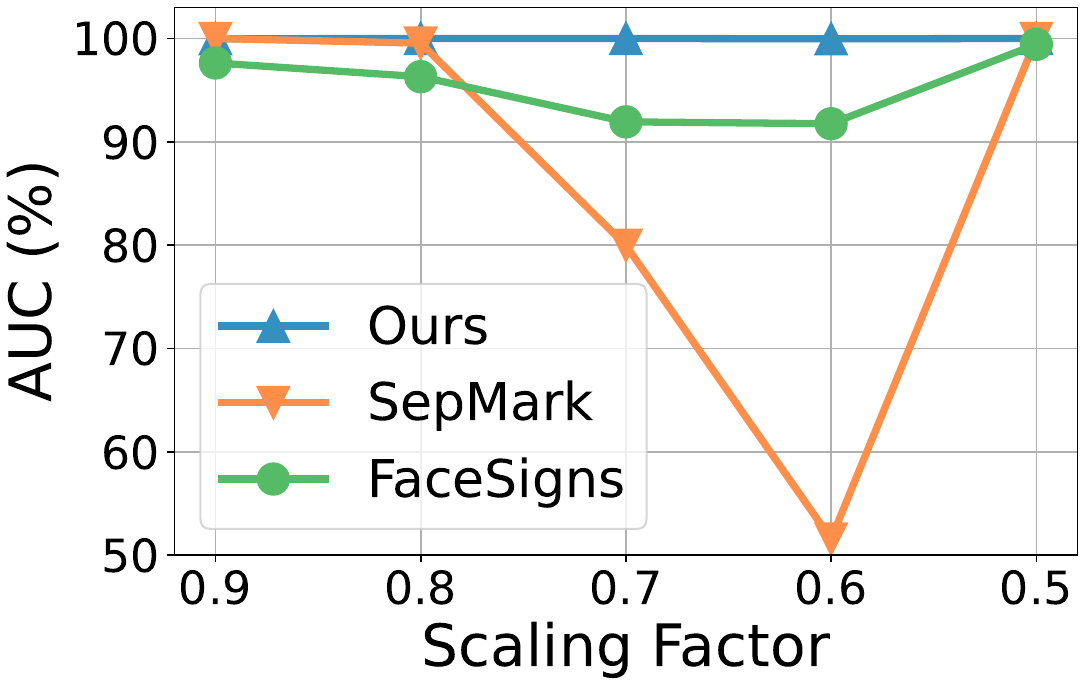}}
  \hfil
  \subfloat[Gaussian Blur]{\includegraphics[width=0.22\textwidth]{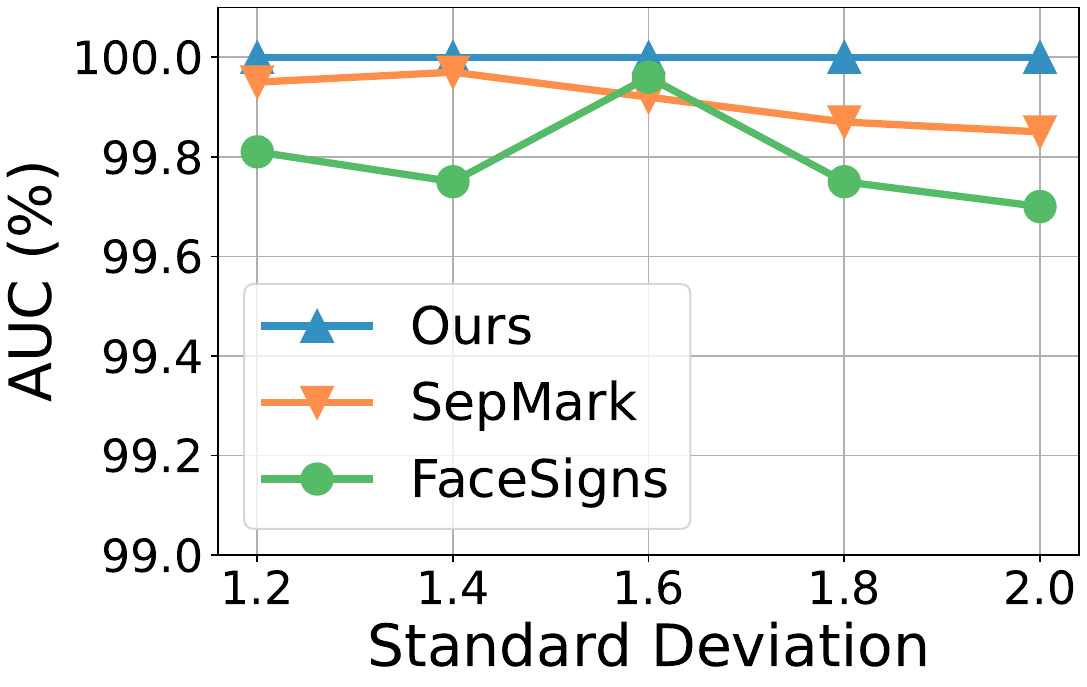}}
  \caption{Robustness evaluation with different benign manipulation intensities on FFHQ. SimSwap~\cite{Chen2020simswap} is applied as the deepfake method. PDDIW~\cite{Zhao2023pddiw} is much worse in this evaluation so is not shown. 
}
  \label{figure:robust_plot}
\end{figure}

\subsection{Proactive Detection Robustness}
We then evaluate the robustness of the proposed approach 
compared with major proactive detection methods. 
The detection AUCs corresponding to four benign manipulations 
(each configured with a single medium strength)
for CelebA-HQ are reported in Tables~\ref{tab:four-subtables}. 
It is clear that even with image processing applied, 
the proposed method still accurately discriminate 
malicious and benign manipulations, outperforming 
the literature proactive methods. 

We further analyze the detection AUCs corresponding to varying benign manipulation strengths, 
with results shown in Figure~\ref{figure:robust_plot}. 
Herein, the face swapping algorithm SimSwap~\cite{Chen2020simswap} 
is used as the malicious manipulation, for a case study. 
The detection performance of the proposed approach remains 
stable even with benign manipulation becoming stronger. 
In contrast, SepMark~\cite{Wu2023sepmark} and FaceSigns~\cite{neekhara2024FaceSigns} 
suffer from performance degradation to different extents. 
\begin{table}[t]
\centering
\footnotesize
\tabcolsep=0.5em
\begin{tabular}{ccccc}
\toprule
Dataset                   & Method   & PSNR $\uparrow$                 & SSIM $\uparrow$                 & LPIPS $\downarrow$               \\ \midrule
\multirow{4}{*}{FFHQ}      & SepMark   & 38.02                & 0.924               & 0.047               \\
                           & FaceSigns & 37.73                & 0.925               & 0.050               \\
                           & PDDIW     & 27.72                & 0.827               & 0.158               \\
                           \cmidrule{2-5}
                           & Ours      & \textbf{56.94}                & \textbf{0.997}               & \textbf{0.001}               \\ \midrule
\multirow{4}{*}{CelebA-HQ} & SepMark   & 38.82                & 0.934               & 0.045               \\
                           & FaceSigns &  39.06                    &    0.933                  &     0.034                 \\
                           & PDDIW     &   28.58                   &     0.843                 &     0.158                 \\
                           \cmidrule{2-5}
                           & Ours      & \textbf{55.82}                & \textbf{0.998}               & \textbf{0.001}               \\ \midrule
\multirow{4}{*}{VGGFace2}  & SepMark   & 41.67                & 0.950               & 0.051               \\
                           & FaceSigns & 39.34 & 0.943 & 0.037 \\
                           & PDDIW     & 30.81                & 0.879               & 0.211               \\
                           \cmidrule{2-5}
                           & Ours      &      \textbf{57.99}                &     \textbf{0.997}                &          \textbf{0.002}            \\ \bottomrule
\end{tabular}%
\caption{Visual quality of steganographic images compared to major proactive methods on different datasets.}
\label{table:visual_quality}
\end{table}

\subsection{Image Quality Evaluation} 
The steganographic image quality of the proposed approach 
in comparison with the other three proactive methods is shown in Table~\ref{table:visual_quality}. 
It is evident that the proposed approach outperforms the other methods with significant margins 
on all three quality metrics. 
In particular, PDDIW~\cite{Zhao2023pddiw} fuses watermarks within facial identity attributes, 
leading to visible textural artifacts in the steganographic image. 
The quality of SepMark~\cite{Wu2023sepmark} and FaceSigns~\cite{neekhara2024FaceSigns} 
remain perceptually high, at the cost of detection accuracy and robustness.

\subsection{Ablation Study and Discussions} 
Last but not least, we verify the effectiveness of each distinctive component. 
As the results in Table~\ref{table:ablation} show, 
the detection accuracy declines slightly by removing $\mathcal{L}_{Rec}$, 
showing the effectiveness of restoring the template image 
in case without malicious manipulation applied. 
If placing no constraint on the face template (w/o $\mathcal{L}_{Sec}$), 
the detector cannot effectively converge. 
This apparently increases the steganographic image quality since the model tends to 
learn a less informative template image that is purely gray. 
By eliminating the benign manipulations (w/o $\mathcal{G}^+$) in training, 
the robustness of deepfake detection cannot be guaranteed 
causing serve decrease in the detection accuracy. 
Without SBI~\cite{Shiohara2022SBI} but using specific manipulations (StarGAN2 and FIT) instead, 
the detection generality is greatly affected. 
Finally, fully utilizing the proposed configurations, 
we achieve a good balance between detection accuracy and image quality. 

\begin{figure}[t]
     \centering
     \includegraphics[width=\columnwidth]{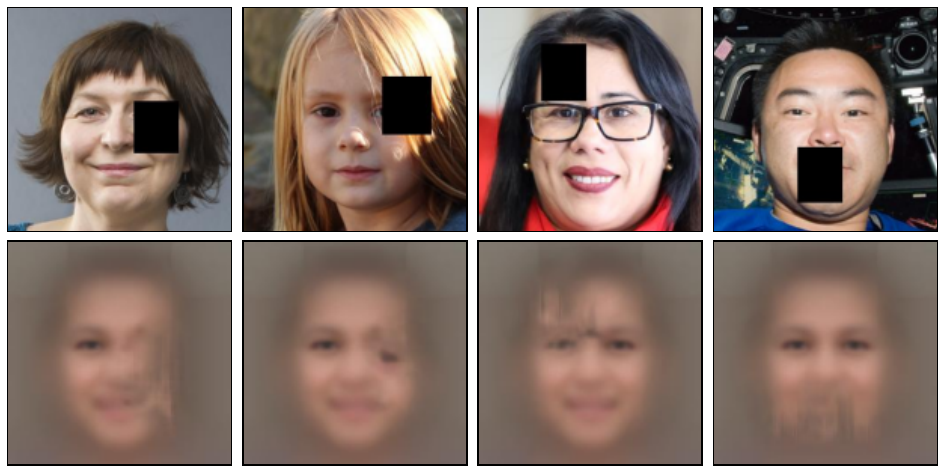}
     \caption{Masked steganographic images (1st row) and corresponding restored templates (2nd row).
     }
     \label{fig:visualization_show_mask}
\end{figure}

To further explain the effectiveness of the proposed approach, 
we randomly mask some regions of the steganographic image, and obtain the corresponding restored face template. 
As the samples in Figure~\ref{fig:visualization_show_mask} show, 
the locations of masked image regions are perfectly reflected 
on the restored templates with visible artifacts. 
This qualitatively demonstrates how the restored template 
can be used for effective deepfake detection.

\begin{table}[t]
\centering
\footnotesize
\tabcolsep=0.3em
\begin{tabular}{lcccccc}
\toprule
\multirow{2}{*}{Config.} & \multicolumn{3}{c}{Detection AUC}  & \multicolumn{3}{c}{Image Quality} \\ \cmidrule(lr){2-4} \cmidrule(lr){5-7}
     & FaceSwap & StarGAN2 & FIT & PSNR & SSIM & LPIPS \\ \midrule 
w/o $\mathcal{L}_{Rec}$            &      97.37       &   99.74      &   99.73  &   62.81   &  0.9998    &    0.0004    \\
w/o $\mathcal{L}_{Sec}$            &      50.00       &    50.00      &  50.00   &   \textbf{71.06}   &   \textbf{0.9999}   &  \textbf{0.0001}      \\
w/o $\mathcal{G}^+$         &      50.00       &      50.00    &   50.00  &   66.85   &   0.9999   &  0.0001      \\
w/o SBI            &      50.09        &   \textbf{100.00}    &   \textbf{100.00}   &   55.79  &  0.9951 &  0.0016   \\ \midrule
Full   &     \textbf{99.96}       &    99.93      &  99.94  &   56.94   &   0.9970   &    0.0011    \\ \bottomrule
\end{tabular}%
\caption{Results of ablation study conducted with JPEG (QF=75) as benign manipulation on FFHQ.}
\label{table:ablation}
\end{table}

\section{Conclusion}
In this paper, we propose a novel proactive deepfake detection framework 
based on the concept of ``hiding a learnable face within a face''. 
Relying on a semi-fragile invertible steganography network, 
a secret template is embedded into a host image imperceptibly, 
acting as a indicator monitoring for any malicious image forgery 
when being restored by the inverse steganography process. 
The secret template is optimized during training to resemble a neutral facial appearance, 
just like a ``big brother'' hidden in the image to be protected. 
By incorporating the self-blending mechanism and robustness learning 
into a simulative transmission channel, a deepfake detector is built to 
accurately distinguish if the steganographic image is maliciously tampered or benignly processed. 
Extensive experiments conducted on multiple datasets demonstrate 
the superiority of the proposed approach over competing literature passive and proactive methods. 

\bibliography{sections/aaai25}
\clearpage
\end{document}


\maketitle

\subsubsection{Appendix A: Additional Detection Robustness Results}
\begin{figure}[H]
  \centering
  \subfloat[JPEG]{\includegraphics[width=0.22\textwidth]{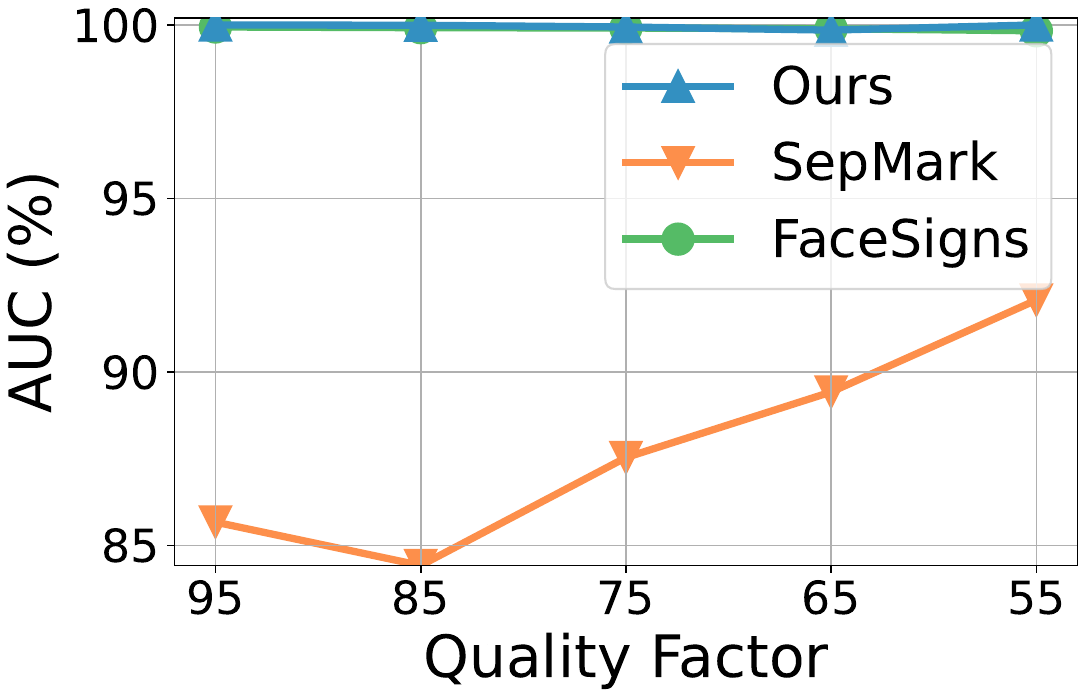}}
  \hfil
  \subfloat[Gaussian Noise]{\includegraphics[width=0.22\textwidth]{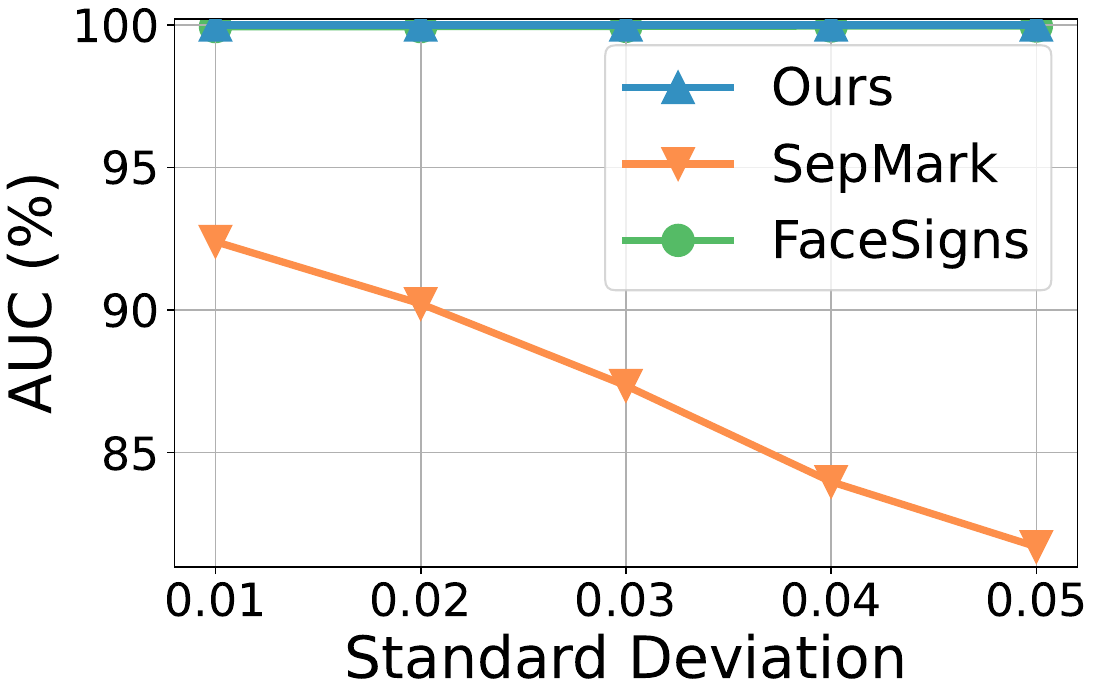}}
  \\
  \subfloat[Resize]{\includegraphics[width=0.22\textwidth]{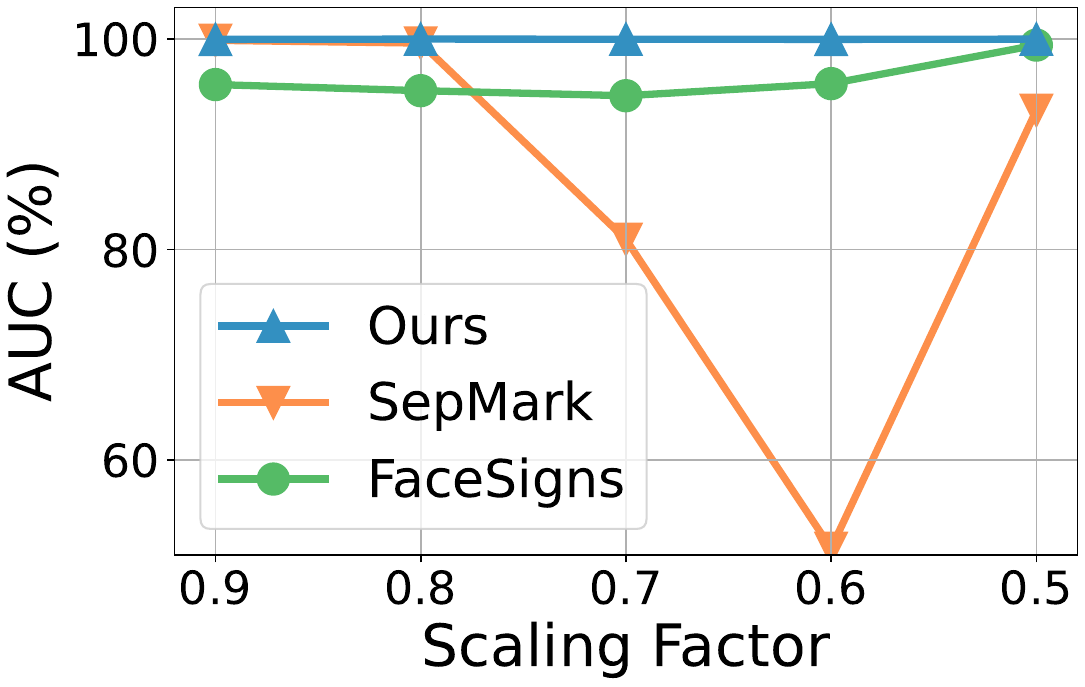}}
  \hfil
  \subfloat[Gaussian Blur]{\includegraphics[width=0.22\textwidth]{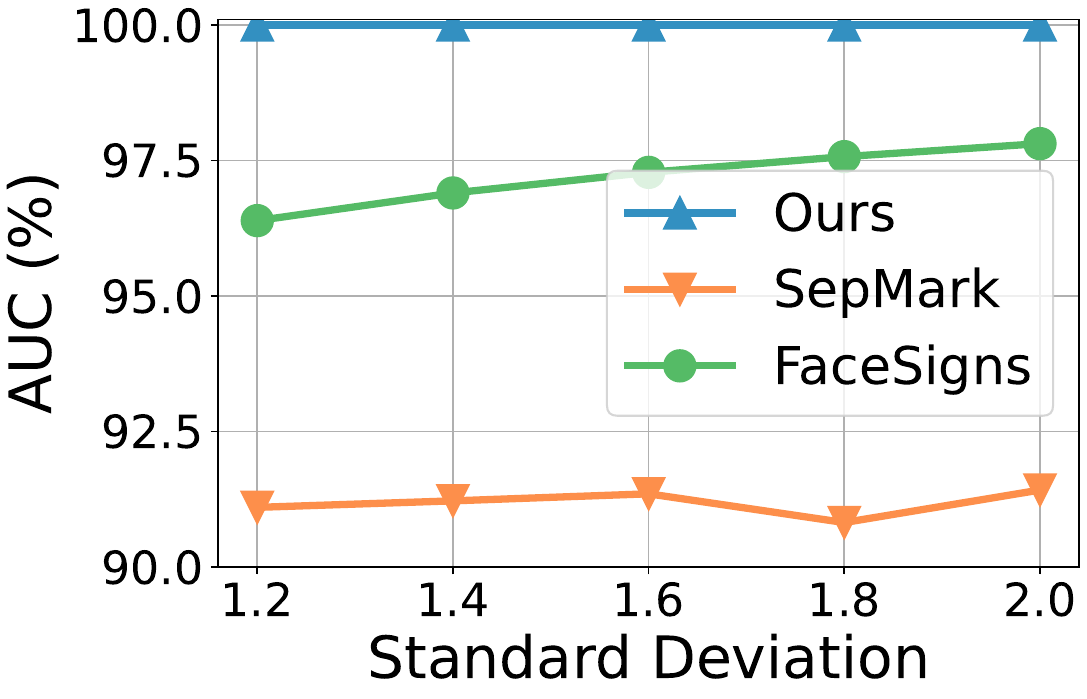}}
  \caption{Robustness evaluation with different benign manipulation intensities on FFHQ. \textbf{FIT} is applied as the deepfake method. PDDIW is much worse in this evaluation so is not shown. 
}
  \label{figure:robust_plot_fit}
\end{figure}

\begin{figure}[H]
  \centering
  \subfloat[JPEG]{\includegraphics[width=0.22\textwidth]{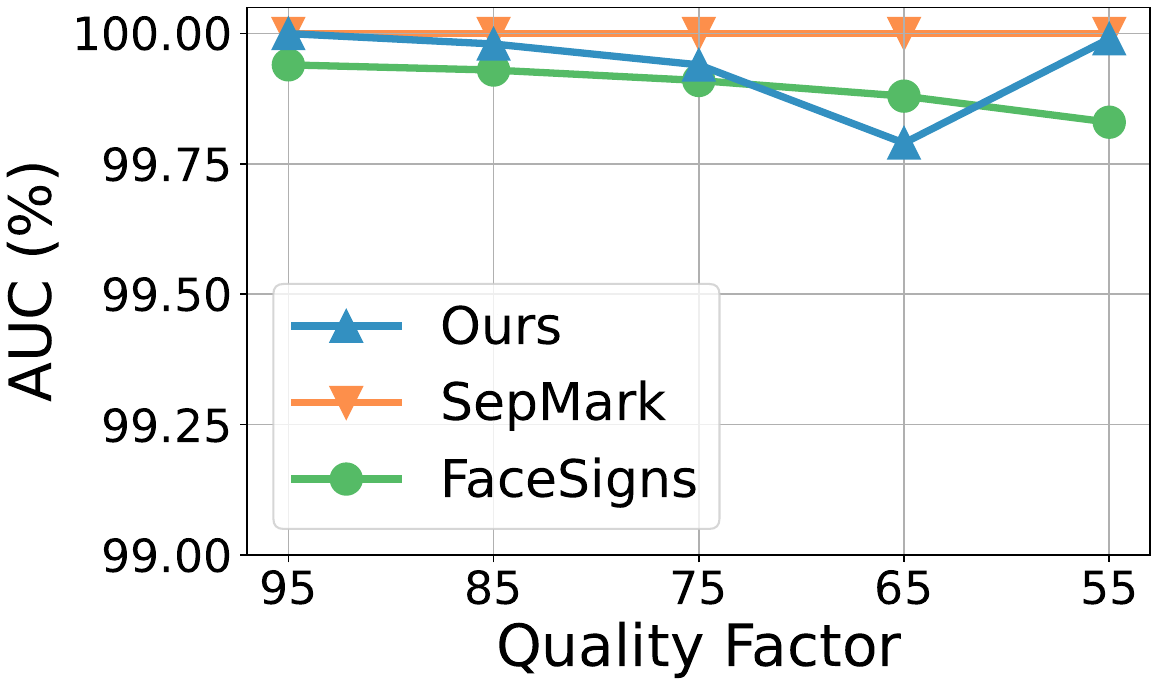}}
  \hfil
  \subfloat[Gaussian Noise]{\includegraphics[width=0.22\textwidth]{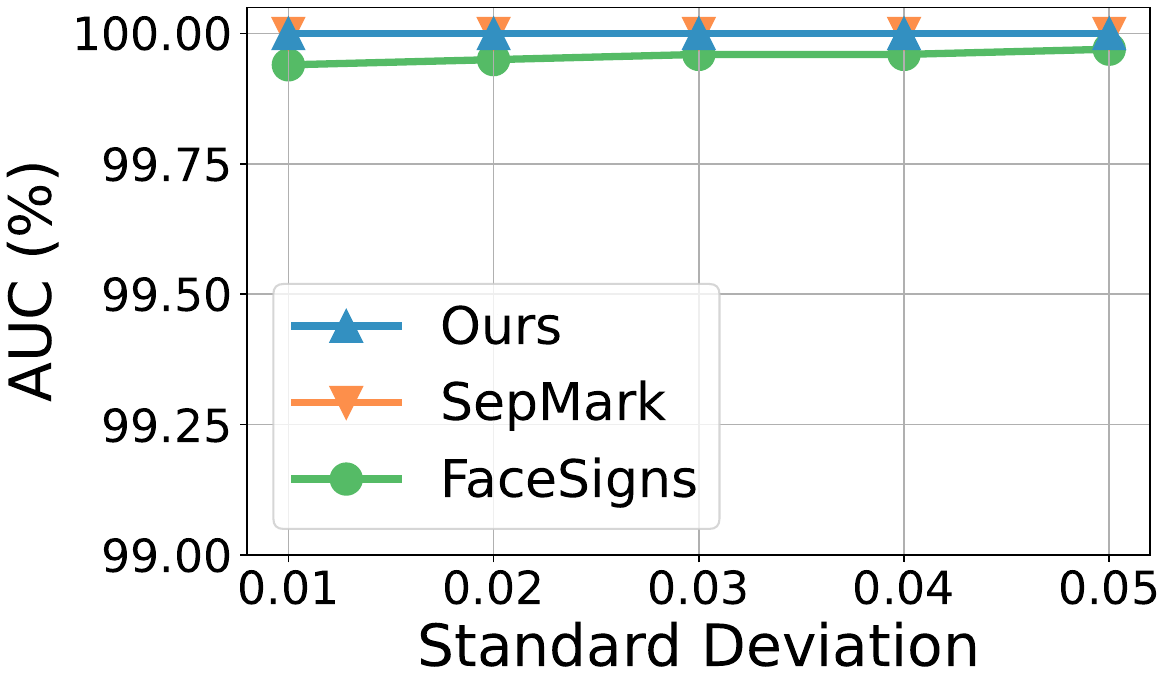}}
  \\
  \subfloat[Resize]{\includegraphics[width=0.22\textwidth]{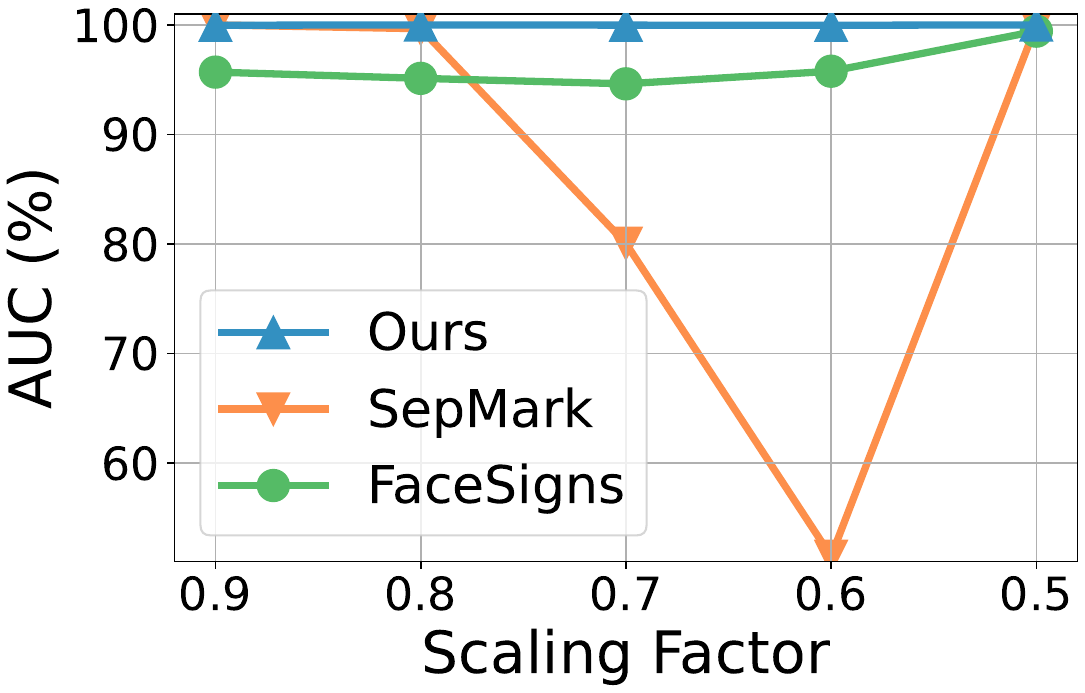}}
  \hfil
  \subfloat[Gaussian Blur]{\includegraphics[width=0.22\textwidth]{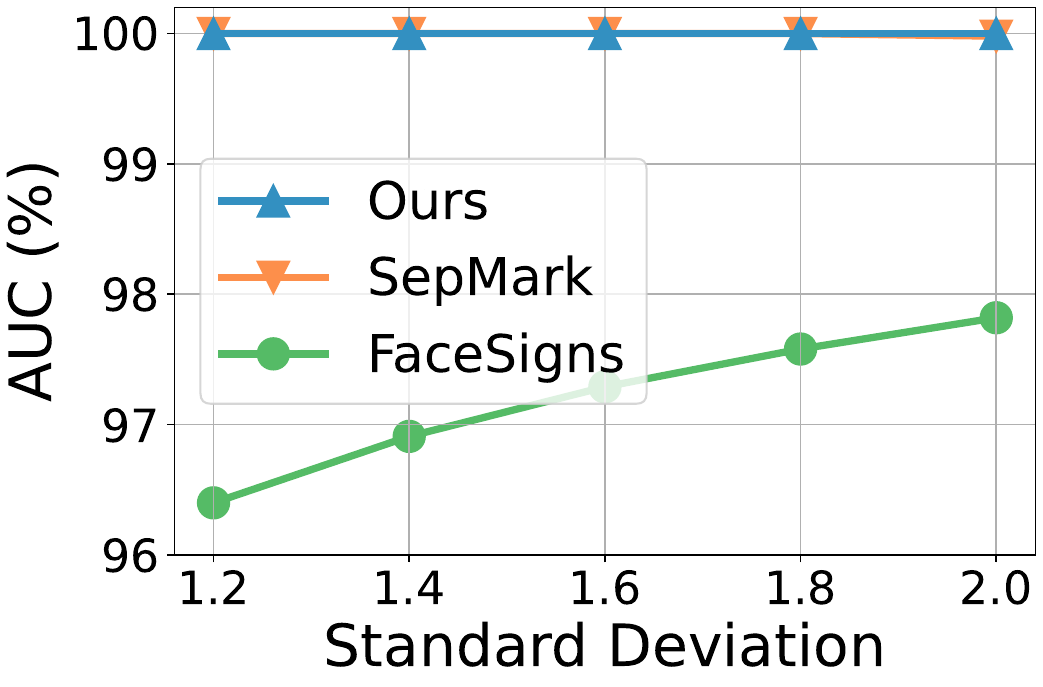}}
  \caption{Robustness evaluation with different benign manipulation intensities on FFHQ. \textbf{StarGAN2} is applied as the deepfake method. PDDIW is much worse in this evaluation so is not shown. 
}
  \label{figure:robust_plot_stargan2}
\end{figure}

As shown in Table~\ref{tab:robust_eval_ffhq} and Table~\ref{tab:robust_eval_vggface2}, 
we further evaluate the robustness of detection performance on FFHQ~\cite{Karras2019ffhq} and VGGFace2~\cite{cao2018vggface2} datasets. 
Our method still achieves the highest average detection AUC against various malicious manipulations when compared to SepMark~\cite{Wu2023sepmark}, FaceSigns~\cite{neekhara2024FaceSigns}, and PDDIW~\cite{Zhao2023pddiw}.
Additionally, as depicted in Figure~\ref{figure:robust_plot_fit} and Figure~\ref{figure:robust_plot_stargan2}, 
with another two deepfake methods (FIT~\cite{gu2019fit} and StartGAN2~\cite{Choi2020starganv2}) 
in varying strengths applied,
our detection performance does not degrade significantly. 
This further verifies the robustness of the proposed approach.

\subsubsection{Appendix B: Additional Qualitative Samples}
\begin{figure*}[t]
     \centering
     \includegraphics[width=0.98\textwidth]{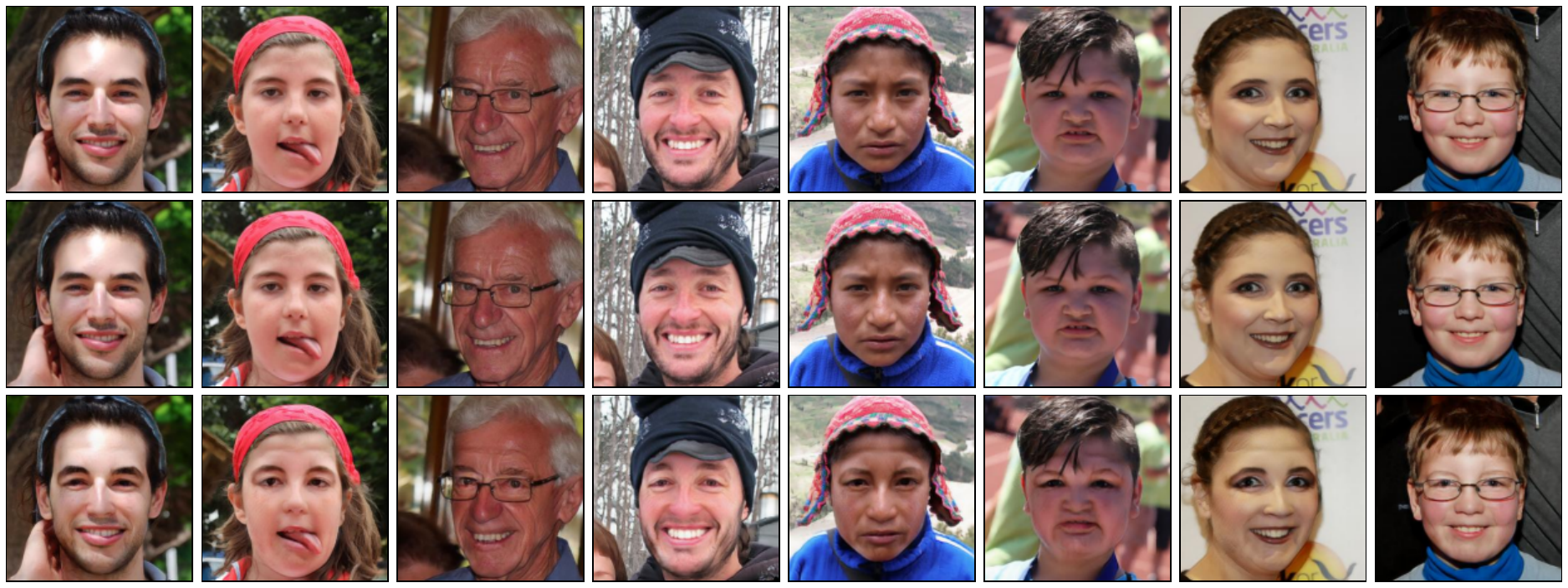}
     \caption{Steganographic image compared with PDDIW on FFHQ.}
     \label{fig:steganographic_compare_pddiw}
\end{figure*}

\begin{figure*}[t]
     \centering
     \includegraphics[width=0.98\textwidth]{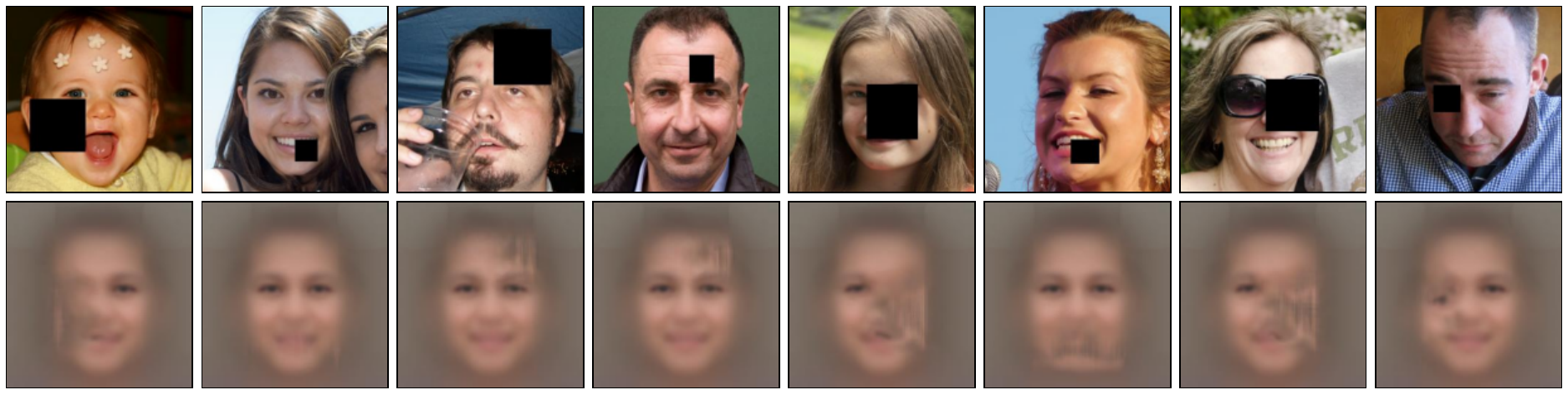}
     \caption{Masked steganographic images (1st row) and corresponding restored templates (2nd row).}
     \label{fig:fig6_visual_mask}
\end{figure*}

\begin{figure*}[t]
     \centering
     \includegraphics[width=0.78\textwidth]{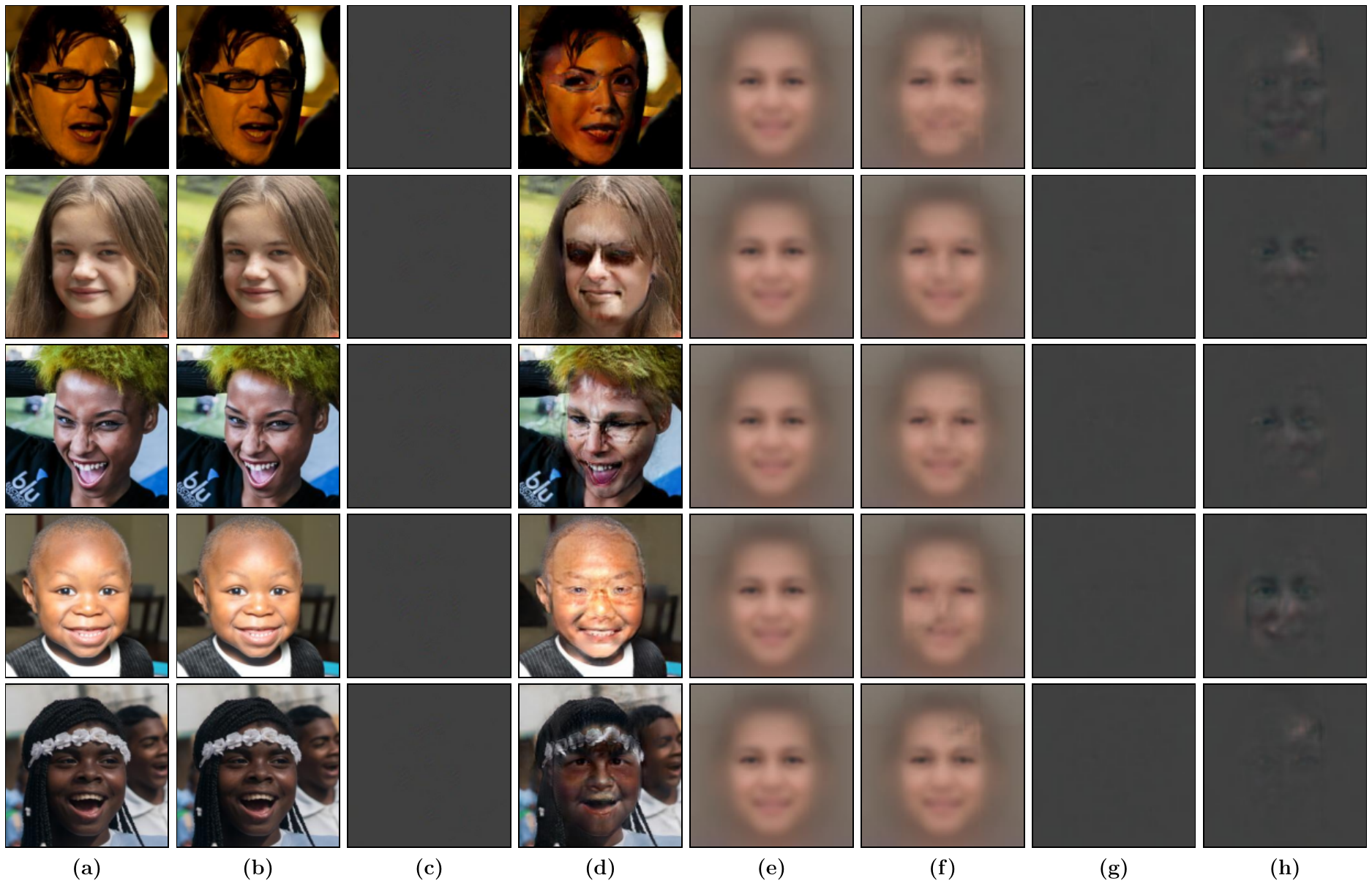}
     \caption{
      Image samples within the framework: 
     (a) Original image; 
     (b) Steganographic image; 
     (c) Residual between (a) and (b); 
     (d) Steganographic image maliciously manipulated by \textbf{FaceShifter}; 
     (e) Template restored from benignly manipulated steganographic image; 
     (f) Template restored from maliciously manipulated steganographic image; 
     (g) Residual between (e) and the original template;
     (h) Residual between (f) and the original template. 
     All above residual images are amplified by a nonlinear square operation for more visible display. 
     }
     \label{fig:visualization_show_all_faceshifter}
\end{figure*}

\begin{figure*}[t]
     \centering
     \includegraphics[width=0.78\textwidth]{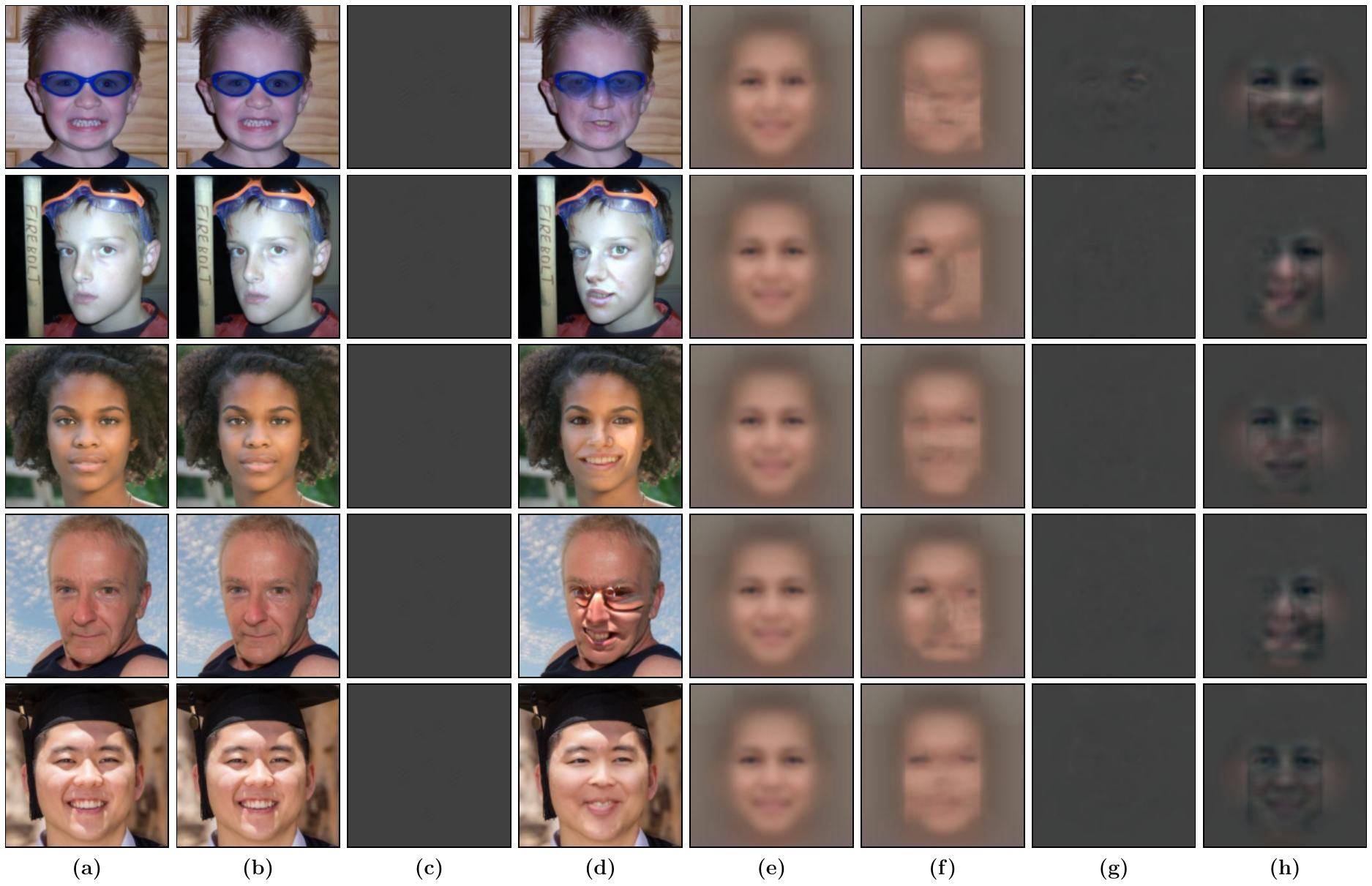}
     \caption{
      Image samples within the framework: 
     (a) Original image; 
     (b) Steganographic image; 
     (c) Residual between (a) and (b); 
     (d) Steganographic image maliciously manipulated by \textbf{FaceSwap}; 
     (e) Template restored from benignly manipulated steganographic image; 
     (f) Template restored from maliciously manipulated steganographic image; 
     (g) Residual between (e) and the original template;
     (h) Residual between (f) and the original template. 
     All above residual images are amplified by a nonlinear square operation for more visible display. 
     }
     \label{fig:visualization_show_all_faceswap}
\end{figure*}

\begin{figure*}[t]
     \centering
     \includegraphics[width=0.78\textwidth]{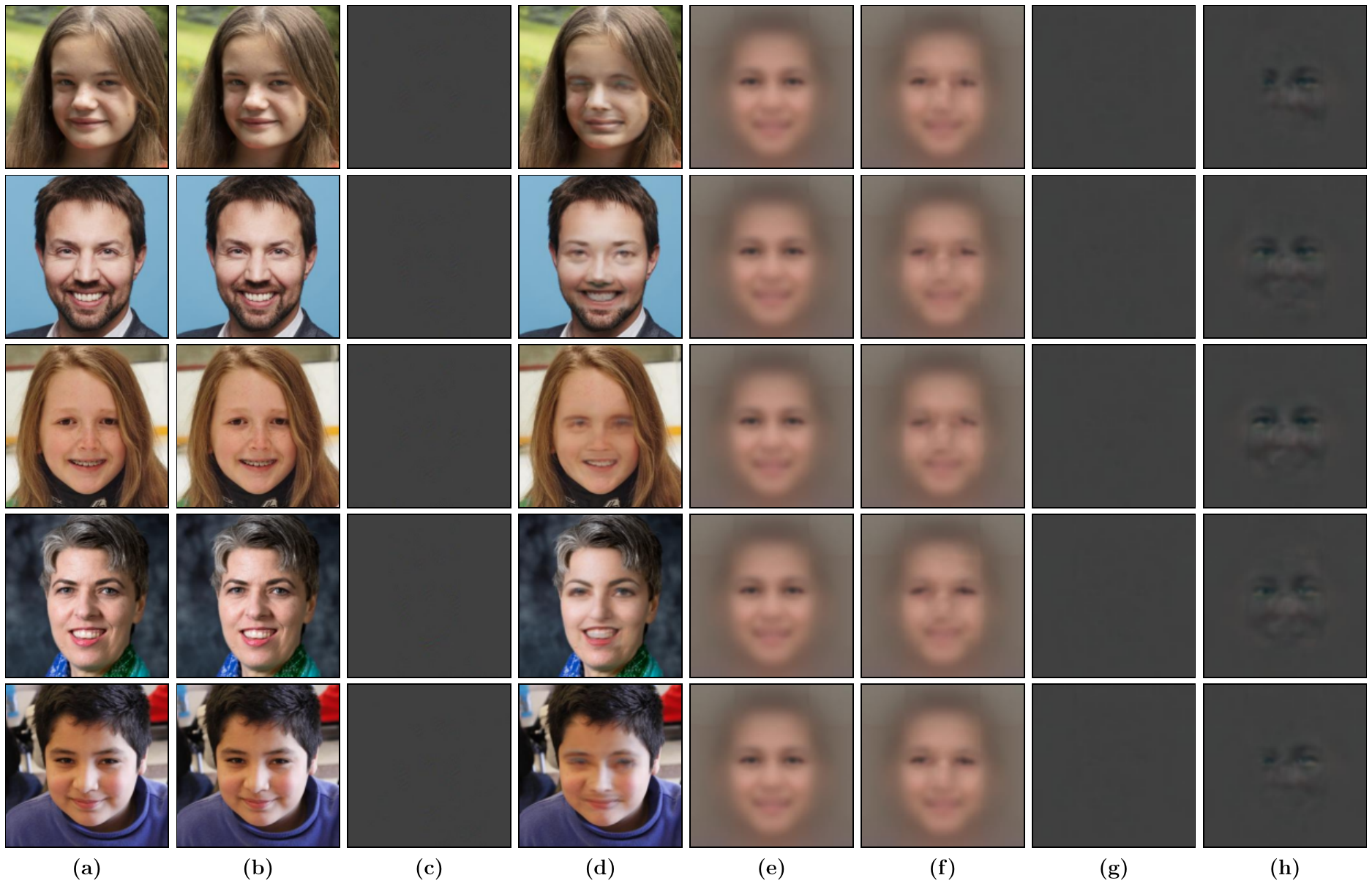}
     \caption{
      Image samples within the framework: 
     (a) Original image; 
     (b) Steganographic image; 
     (c) Residual between (a) and (b); 
     (d) Steganographic image maliciously manipulated by \textbf{MFaceSwap}; 
     (e) Template restored from benignly manipulated steganographic image; 
     (f) Template restored from maliciously manipulated steganographic image; 
     (g) Residual between (e) and the original template;
     (h) Residual between (f) and the original template. 
     All above residual images are amplified by a nonlinear square operation for more visible display. 
     }
     \label{fig:visualization_show_all_mobileswap}
\end{figure*}

\begin{figure*}[t]
     \centering
     \includegraphics[width=0.78\textwidth]{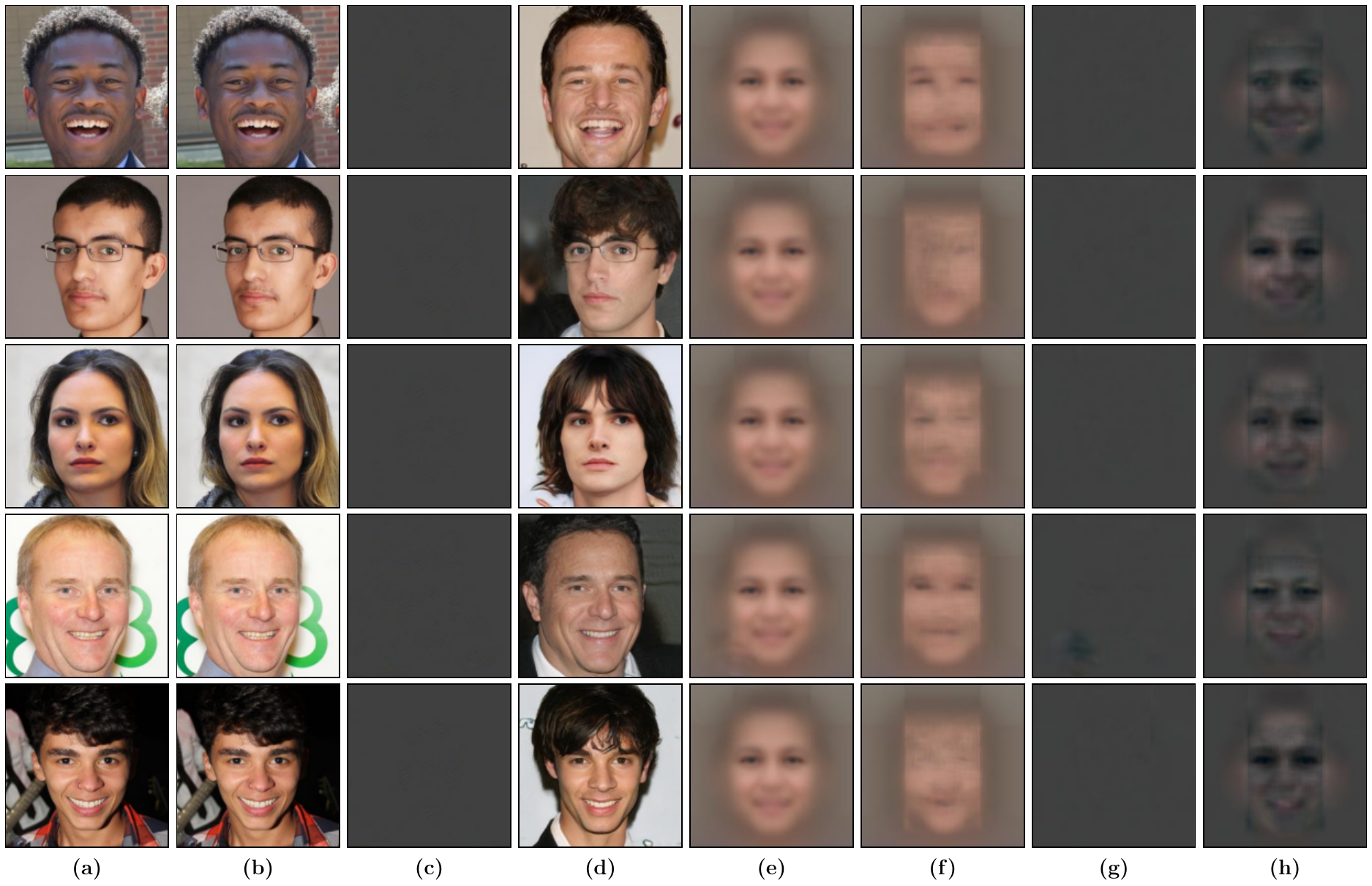}
     \caption{
      Image samples within the framework: 
     (a) Original image; 
     (b) Steganographic image; 
     (c) Residual between (a) and (b); 
     (d) Steganographic image maliciously manipulated by \textbf{StarGAN2}; 
     (e) Template restored from benignly manipulated steganographic image; 
     (f) Template restored from maliciously manipulated steganographic image; 
     (g) Residual between (e) and the original template;
     (h) Residual between (f) and the original template. 
     All above residual images are amplified by a nonlinear square operation for more visible display. 
     }
     \label{fig:visualization_show_all_stargan2}
\end{figure*}

\begin{figure*}[t]
     \centering
     \includegraphics[width=0.78\textwidth]{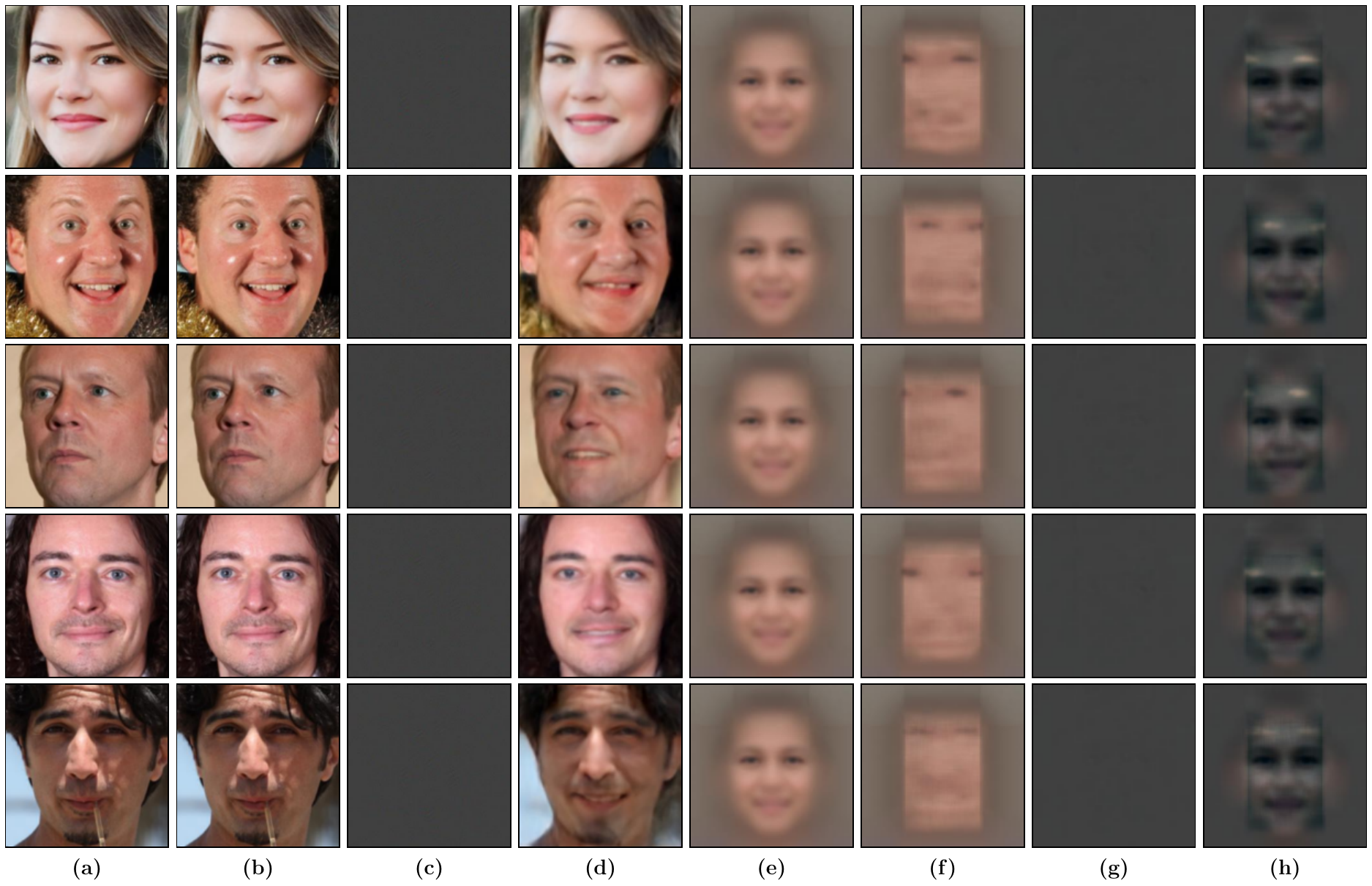}
     \caption{
      Image samples within the framework: 
     (a) Original image; 
     (b) Steganographic image; 
     (c) Residual between (a) and (b); 
     (d) Steganographic image maliciously manipulated by \textbf{TTedit}; 
     (e) Template restored from benignly manipulated steganographic image; 
     (f) Template restored from maliciously manipulated steganographic image; 
     (g) Residual between (e) and the original template;
     (h) Residual between (f) and the original template. 
     All above residual images are amplified by a nonlinear square operation for more visible display. 
     }
     \label{fig:visualization_show_all_ttedit}
\end{figure*}

\begin{figure*}[t]
     \centering
     \includegraphics[width=0.78\textwidth]{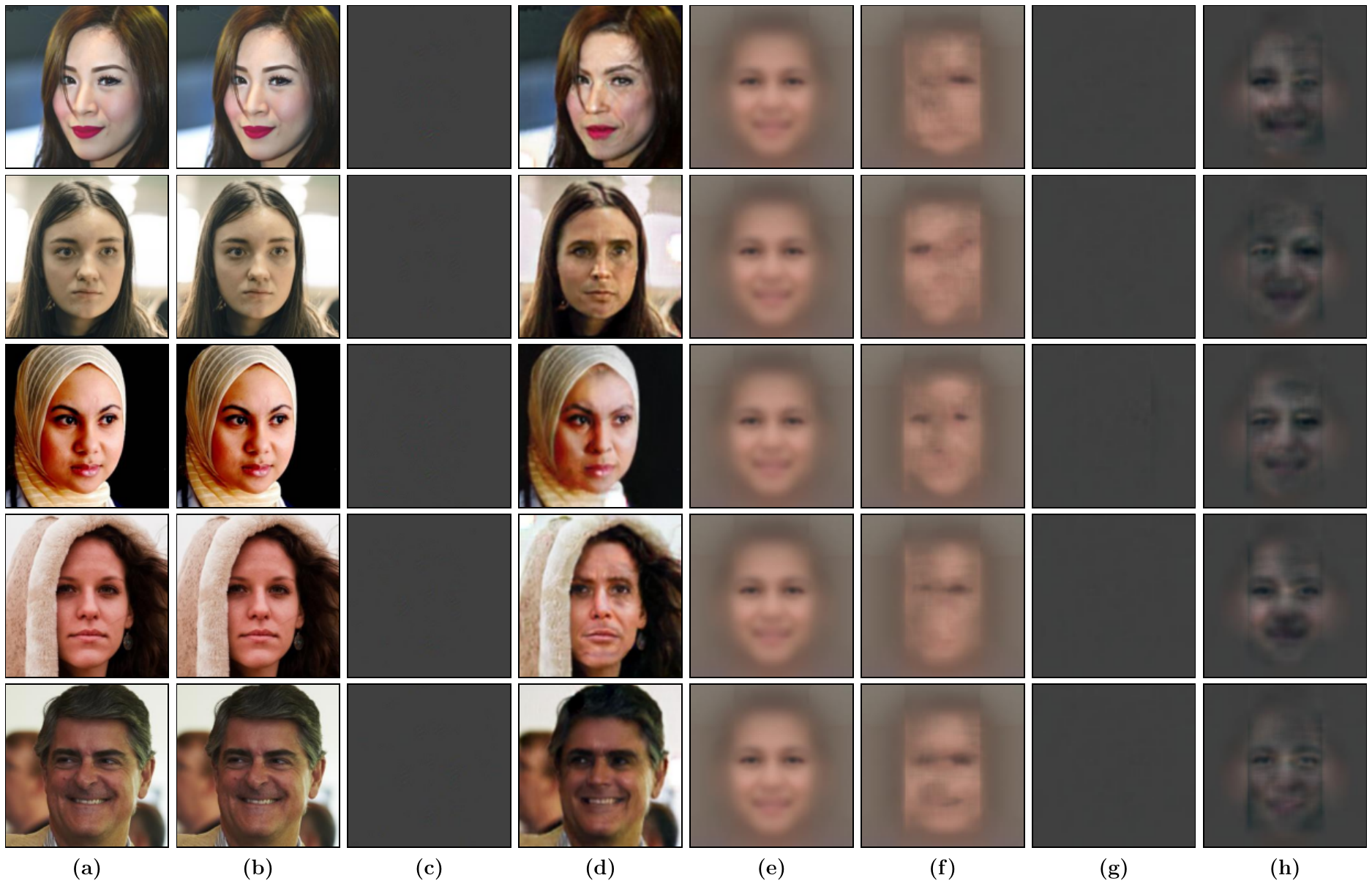}
     \caption{
      Image samples within the framework: 
     (a) Original image; 
     (b) Steganographic image; 
     (c) Residual between (a) and (b); 
     (d) Steganographic image maliciously manipulated by \textbf{FIT}; 
     (e) Template restored from benignly manipulated steganographic image; 
     (f) Template restored from maliciously manipulated steganographic image; 
     (g) Residual between (e) and the original template;
     (h) Residual between (f) and the original template. 
     All above residual images are amplified by a nonlinear square operation for more visible display. 
     }
     \label{fig:visualization_show_all_fit}
\end{figure*}

Figure~\ref{fig:steganographic_compare_pddiw} compares the steganographic images 
between the proposed method and PDDIW~\cite{Zhao2023pddiw}. It is clear that the proposed method generates 
almost imperceptible steganographic results whereas PDDIW~\cite{Zhao2023pddiw} introduces slight changes 
to the steganographic images when compared to the original. 
Figure~\ref{fig:fig6_visual_mask} shows more samples of regional-masked steganographic images 
and their corresponding restored secret templates, demonstrating the correspondence 
between masking traces and restored template artifacts in the same masking locations. 
Moreover, in Figures~\ref{fig:visualization_show_all_faceshifter}-
\ref{fig:visualization_show_all_fit}, we provide more samples of 
steganographic results (and their residuals to the original image), 
deepfake results, restored image template, and the template image residuals, 
obtained for multiple deepfake operations.

\begin{table}[H]
\centering
\tabcolsep=0.3em
\begin{tabular}{c}
    \begin{minipage}{\linewidth}
        \centering
        \resizebox{\linewidth}{!}{%
        \begin{tabular}{lccccccc}
        \toprule
        \multirow{2}{*}{\begin{tabular}[c]{@{}c@{}}\bf JPEG\\ \bf (QF=75)\end{tabular}} & \multicolumn{4}{c}{Identity Swap} & \multicolumn{2}{c}{Attribute Edit} & FA \\ \cmidrule(lr){2-5} \cmidrule(lr){6-7} \cmidrule(lr){8-8} 
                                & SS & FST & FS & MFS & SG2 & TTE & FIT \\
        \midrule 
        PDDIW & 96.90 & 98.91 & 63.13 & 85.95 & 85.68 & 54.15 & 99.00 \\
        SepMark & 99.05 & 50.00 & 50.00 & 50.00 & \textbf{100.00} & 81.68 & 86.60 \\
        FaceSigns & 99.67 & 99.83 & 99.87 & 99.53 & 99.95 & 99.90 & \textbf{99.95} \\
        \midrule
        Ours & \textbf{99.94} & \textbf{99.96} & \textbf{99.91} & \textbf{99.78} & 99.93 & \textbf{99.90} & 99.94 \\
        \bottomrule
        \end{tabular}
        }
    \end{minipage}%
    
    \\
    \addlinespace 

    \begin{minipage}{\linewidth} 
        \centering
        \resizebox{\linewidth}{!}{%
        \begin{tabular}{lccccccc}
        \toprule
         \multirow{2}{*}{\begin{tabular}[c]{@{}c@{}}\bf Noising \\ \textbf{(\bm{$\sigma$}=0.03)} \end{tabular}}   & \multicolumn{4}{c}{Identity Swap}  & \multicolumn{2}{c}{Attribute Edit} & FA \\ \cmidrule(lr){2-5} \cmidrule(lr){6-7} \cmidrule(lr){8-8} 
                                & SS & FST & FS & MFS & SG2 & TTE & FIT \\
        \midrule
        PDDIW & 97.03 & 98.93 & 63.97 & 86.87 & 86.20 & 53.65 & 99.03 \\
        SepMark & 98.73 & 50.00 & 50.00 & 50.00 & 100.00 & 99.04 & 87.07 \\
        FaceSigns & 99.58 & 99.90 & 99.95 & 99.68 & 99.98 & 98.95 & 100.00 \\
        \midrule
        Ours & \textbf{100.00} & \textbf{100.00} & \textbf{100.00} & \textbf{99.79} & \textbf{100.00} & \textbf{99.50} & \textbf{100.00} \\
        \bottomrule
        \end{tabular}
        }
    \end{minipage}

    \\
    \addlinespace 
    
    \begin{minipage}{\linewidth} 
        \centering
        \resizebox{\linewidth}{!}{
        \begin{tabular}{lccccccc}
        \toprule
        \multirow{2}{*}{\begin{tabular}[c]{@{}c@{}}\bf Rescaling\\ \bf (Factor=0.7)\end{tabular}} & \multicolumn{4}{c}{Identity Swap} & \multicolumn{2}{c}{Attribute Edit} & FA \\ \cmidrule(lr){2-5} \cmidrule(lr){6-7} \cmidrule(lr){8-8} 
                                & SS & FST & FS & MFS & SG2 & TTE & FIT \\
        \midrule
        PDDIW & 96.87 & 98.83 & 63.35 & 84.90 & 86.08 & 54.37 & 98.97 \\
        SepMark & 79.97 & 79.90 & 74.48 & 78.58 & 80.68 & 98.25 & 80.90 \\
        FaceSigns & 91.93 & 92.27 & 92.48 & 92.00 & 92.33 & 96.25 & 92.33 \\
        \midrule
        Ours & \textbf{100.00} & \textbf{99.99} & \textbf{99.96} & \textbf{99.81} & \textbf{99.97} & \textbf{99.81} & \textbf{100.00} \\
        \bottomrule
        \end{tabular}
        }
    \end{minipage}%
    
    \\
    \addlinespace 

    \begin{minipage}{\linewidth} 
        \centering
        \resizebox{\linewidth}{!}{%
        \begin{tabular}{lccccccc}
        \toprule 
        \multirow{2}{*}{\begin{tabular}[c]{@{}c@{}}\bf Blur \\ \textbf{(\bm{$\sigma$}=1.5)} \end{tabular}} & \multicolumn{4}{c}{Identity Swap} & \multicolumn{2}{c}{Attribute Edit} & FA \\ \cmidrule(lr){2-5} \cmidrule(lr){6-7} \cmidrule(lr){8-8} 
                                & SS & FST & FS & MFS & SG2 & TTE & FIT \\
        \midrule
        PDDIW & 96.73 & 98.87 & 63.48 & 86.28 & 85.72 & 53.79 & 98.95 \\
        SepMark & 99.92 & 50.00 & 50.00 & 50.00 & 100.00 & 99.96 & 91.23 \\
        FaceSigns & 99.75 & 99.88 & 99.93 & 99.68 & 99.95 & \textbf{100.00} & 99.93 \\
        \midrule
        Ours & \textbf{100.00} & \textbf{100.00} & \textbf{100.00} & \textbf{99.82} & \textbf{100.00} & 99.96 & \textbf{100.00} \\
        \bottomrule
        \end{tabular}
        }
    \end{minipage}

\end{tabular}
\caption{Robustness evaluation with different benign transformations on \textbf{FFHQ}. SS, FST, FS, MFS, SG2, TTE, and FA refer to SimSwap, FaceShifter, FaceSwap, MobileFaceSwap, StarGAN2, TTedit, and face anonymization, respectively.}
\label{tab:robust_eval_ffhq}
\end{table}

\begin{table}[H]
\centering
\tabcolsep=0.3em
\begin{tabular}{c}

    \begin{minipage}{\linewidth}
        \centering
        \resizebox{\linewidth}{!}{%
        \begin{tabular}{lccccccc}
        \toprule
        \multirow{2}{*}{\begin{tabular}[c]{@{}c@{}}\bf JPEG\\ \bf (QF=75)\end{tabular}} & \multicolumn{4}{c}{Identity Swap} & \multicolumn{2}{c}{Attribute Edit} & FA \\ \cmidrule(lr){2-5} \cmidrule(lr){6-7} \cmidrule(lr){8-8} 
                                & SS & FST & FS & MFS & SG2 & TTE & FIT \\
        \midrule 
        PDDIW & 58.23 & 57.82 & 58.37 & 57.83 & 57.23 & 58.27 & 56.32 \\
        SepMark & 93.75 & 50.00 & 50.14 & 50.00 & 100.00 & 100.00 & 59.70 \\
        FaceSigns & 97.25 & 99.20 & 99.48 & 96.08 & 99.75 & 99.90 & 97.65 \\
        \midrule
        Ours & \textbf{100.00} & \textbf{100.00} & \textbf{100.00} & \textbf{99.66} & \textbf{100.00} & \textbf{100.00} & \textbf{100.00} \\
        \bottomrule
        \end{tabular}
        }
    \end{minipage}%
    
    \\
    \addlinespace 

    \begin{minipage}{\linewidth} 
        \centering
        \resizebox{\linewidth}{!}{%
        \begin{tabular}{lccccccc}
        \toprule
         \multirow{2}{*}{\begin{tabular}[c]{@{}c@{}}\bf Noising \\ \textbf{(\bm{$\sigma$}=0.03)} \end{tabular}}   & \multicolumn{4}{c}{Identity Swap}  & \multicolumn{2}{c}{Attribute Edit} & FA \\ \cmidrule(lr){2-5} \cmidrule(lr){6-7} \cmidrule(lr){8-8} 
                                & SS & FST & FS & MFS & SG2 & TTE & FIT \\
        \midrule
        PDDIW & 57.27 & 56.88 & 57.10 & 56.97 & 56.32 & 57.29 & 55.62 \\
        SepMark & 99.82 & 50.03 & 50.03 & 50.00 & 100.00 &\textbf{ 100.00} & 98.03 \\
        FaceSigns & 94.73 & 97.73 & 98.40 & 95.80 & 98.30 & 98.30 & 93.05 \\
        \midrule
        Ours & \textbf{100.00} & \textbf{100.00} & \textbf{100.00} & \textbf{99.62} & \textbf{100.00} & 99.55 & \textbf{100.00} \\
        \bottomrule
        \end{tabular}
        }
    \end{minipage}

    \\
    \addlinespace 
    
    \begin{minipage}{\linewidth} 
        \centering
        \resizebox{\linewidth}{!}{
        \begin{tabular}{lccccccc}
        \toprule
        \multirow{2}{*}{\begin{tabular}[c]{@{}c@{}}\bf Rescaling\\ \bf (Factor=0.7)\end{tabular}} & \multicolumn{4}{c}{Identity Swap} & \multicolumn{2}{c}{Attribute Edit} & FA \\ \cmidrule(lr){2-5} \cmidrule(lr){6-7} \cmidrule(lr){8-8} 
                                & SS & FST & FS & MFS & SG2 & TTE & FIT \\
        \midrule
        PDDIW & 57.72 & 57.31 & 57.88 & 57.25 & 58.82 & 57.75 & 55.75 \\
        SepMark & 91.53 & 90.28 & 81.49 & 85.07 & 91.15 & 83.98 & 91.73 \\
        FaceSigns & 96.15 & 98.03 & 98.37 & 95.86 & 98.67 & 98.15 & 93.65 \\
        \midrule
        Ours & \textbf{100.00} & \textbf{100.00} & \textbf{100.00} & \textbf{99.46} & \textbf{100.00} & \textbf{100.00} & \textbf{100.00} \\
        \bottomrule
        \end{tabular}
        }
    \end{minipage}%
    
    \\
    \addlinespace 

    \begin{minipage}{\linewidth} 
        \centering
        \resizebox{\linewidth}{!}{%
        \begin{tabular}{lccccccc}
        \toprule 
        \multirow{2}{*}{\begin{tabular}[c]{@{}c@{}}\bf Blur \\ \textbf{(\bm{$\sigma$}=1.5)} \end{tabular}} & \multicolumn{4}{c}{Identity Swap} & \multicolumn{2}{c}{Attribute Edit} & FA \\ \cmidrule(lr){2-5} \cmidrule(lr){6-7} \cmidrule(lr){8-8} 
                                & SS & FST & FS & MFS & SG2 & TTE & FIT \\
        \midrule
        PDDIW & 57.31 & 58.90 & 57.39 & 56.93 & 56.35 & 57.35 & 55.45 \\
        SepMark & 100.00 & 50.47 & 50.14 & 50.00 & 99.93 & 100.00 & 95.20 \\
        FaceSigns & 97.25 & 99.51 & 99.58 & 97.06 & 99.98 & \textbf{100.00} & 95.10 \\
        \midrule
        Ours & \textbf{100.00} & \textbf{100.00} & \textbf{100.00} & \textbf{99.55} & \textbf{100.00} & 99.77 & \textbf{100.00} \\
        \bottomrule
        \end{tabular}
        }
    \end{minipage}

\end{tabular}
\caption{Robustness evaluation with different benign transformations on \textbf{VGGFace2}. SS, FST, FS, MFS, SG2, TTE, and FA refer to SimSwap, FaceShifter, FaceSwap, MobileFaceSwap, StarGAN2, TTedit, and face anonymization respectively.}
\label{tab:robust_eval_vggface2}
\end{table}

\bibliography{sections/aaai25}